%% file: main.tex
\documentclass[10pt,twocolumn,letterpaper]{article}

\input{3dvheader}

\begin{document}

\title{Dual Mesh Convolutional Networks for Human Shape Correspondence}

\maketitle
\thispagestyle{empty}

\input{sec_abstract}
\input{sec_intro}

\input{sec_related}

\input{sec_method}
\input{sec_experiments}
\input{sec_conclusion}

{\small
\bibliographystyle{ieee_fullname}
\bibliography{jjv,more}
}

\end{document}


\title{Dual Mesh Convolutional Networks for Human Shape Correspondence \\
– Supplementary Material –} 

\maketitle
\thispagestyle{empty}

\section{Network architectures and training}

We trained models for  MoNet~\cite{monti17cvpr}, SplineCNN~\cite{fey18cvpr} and FeaStNet~\cite{verma18cvpr}  using the PyTorch Geometric framework \cite{Fey2019pyg} for the different  evaluation setups. 
We use the same base architecture for primal and dual based methods as given in Figure 5 of the main paper,  where we replace the DualConvMax blocks with the corresponding convolutional blocks. We use SHOT descriptors \cite{tombari10eccv} as input features for MoNet, and XYZ for SplineCNN and FeaStNet. 
While our DualConvMax operator has three weights per filter by construction, due to the regularity of the dual mesh, for the other methods the number of weights is a hyper-parameter. In our experiments,  we use the same number of weights per filter as in the original papers, \ie 8 for MoNet, FeaStNet(--Dual)  and 5 for SplineCNN. 

\tab{remeshed_time} shows the average execution time of a forward pass for a mesh of the re-meshed Faust dataset for the different models  used in our experiments on a Tesla P100 GPU, as well as the number of parameters of these models.
All  methods have a comparable number of parameters and execution times except SplineCNN. 
The latter takes roughly twice longer to complete a forward pass, and possesses roughly twice as many parameters as the other methods. 
For MoNet, the forward pass time does not include the computation of the SHOT descriptors, which are used as input to the model, and take an additional 400 ms to compute per mesh on CPU.

\begin{table}
\centering
 {
    \begin{tabular}{llrc}
        \toprule
        Domain & Method & Time (ms) & \#Prms.\ \\
        \midrule
        \multirow{3}{*}{Primal} & MoNet & $8.9 \pm 0.25$ & 1.4M \\
         & SplineCNN & $17.8 \pm 0.15$ & 3.1M \\
         & FeaStNet & $10.6 \pm 0.17$ & 1.4M \\
        \midrule
        \multirow{3}{*}{Dual} & FeaStConv--Dual & $10.4 \pm 0.14$ & 1.4M \\
         & DualConvMax & $8.6 \pm 0.57$ & 1.3M \\
        \bottomrule
    \end{tabular}
    }
    \caption{Comparison of the number of parameters and average execution time of the forward-pass per mesh for Faust-Remeshed.
    \label{tab:remeshed_time}
}
\end{table}


\section{Additional Visualizations}

\begin{figure*}
	\setlength{\tabcolsep}{9pt}
	\centering
	\begin{tabular}{lc|ccc|cc}
	 & \small\textbf{Faust-Synthetic} & \small\textbf{MoNet} & \small\textbf{SplineCNN} & \small\textbf{FeaStNet} & \small{\textbf{FeaStNet--Dual}} & \small{\textbf{DualConvMax}}
	\\
	 & \small\textbf{Reference} & \small{(SHOT)} & \small{(XYZ)} & \small{(XYZ)} & \small{(XYZ)} & \small{(XYZ)}
	 \\
	{\rotatebox{90}{\small\textbf{$0\%$}}} &
	\includegraphics[height=120pt]{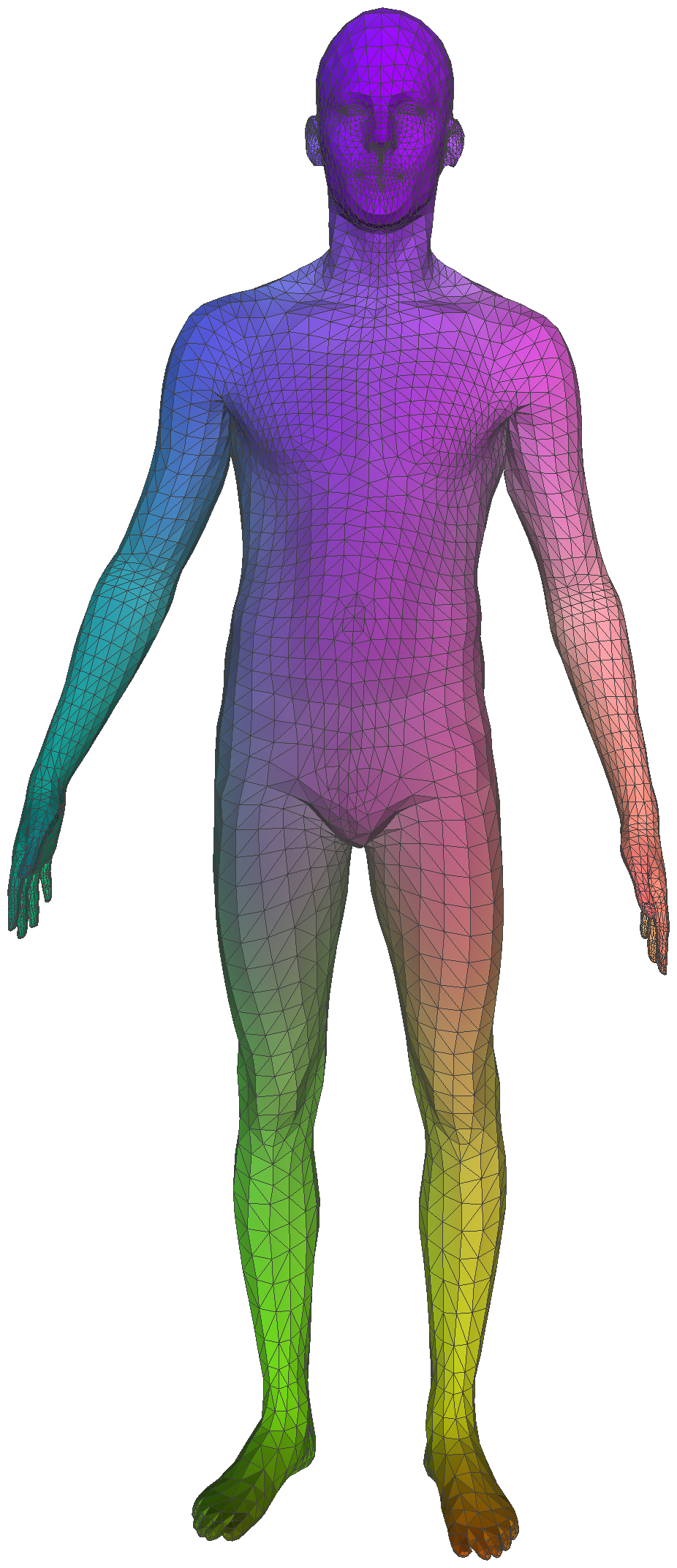} &
	\includegraphics[height=120pt]{faust_decimated_plots/monet_dec_0_train_faust_80} &
	\includegraphics[height=120pt]{faust_decimated_plots/spline_dec_0_train_faust_80} &
	\includegraphics[height=120pt]{faust_decimated_plots/feast_dec_0_train_faust_80} &
	\includegraphics[height=120pt]{faust_decimated_plots/feast_dual_dec_0_train_faust_80} &
	\includegraphics[height=120pt]{faust_decimated_plots/dual_dec_0_train_faust_xyz_80}
	\\
	 &
	\includegraphics[height=60pt]{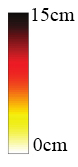} &
	\includegraphics[height=120pt]{faust_decimated_plots/errors/monet_dec_0_train_faust_errors_80} &
	\includegraphics[height=120pt]{faust_decimated_plots/errors/spline_dec_0_train_faust_errors_80} &
	\includegraphics[height=120pt]{faust_decimated_plots/errors/feast_dec_0_train_faust_errors_80} &
	\includegraphics[height=120pt]{faust_decimated_plots/errors/feast_dual_dec_0_train_faust_errors_80} &
	\includegraphics[height=120pt]{faust_decimated_plots/errors/dual_dec_0_train_faust_xyz_errors_80}
	\\ 
	\midrule
	{\rotatebox{90}{\small\textbf{$50\%$}}} &
	\includegraphics[height=120pt]{remeshed_faust_plots/faust_ref}  &
	\includegraphics[height=120pt]{faust_decimated_plots/monet_dec_50_train_faust_80} &
	\includegraphics[height=120pt]{faust_decimated_plots/spline_dec_50_train_faust_80} &
	\includegraphics[height=120pt]{faust_decimated_plots/feast_dec_50_train_faust_80} &
	\includegraphics[height=120pt]{faust_decimated_plots/feast_dual_dec_50_train_faust_80} &
	\includegraphics[height=120pt]{faust_decimated_plots/dual_dec_50_train_faust_xyz_80}
	\\
	&
	\includegraphics[height=60pt]{figures/errors_colormap} &
	\includegraphics[height=120pt]{faust_decimated_plots/errors/monet_dec_50_train_faust_errors_80} &
	\includegraphics[height=120pt]{faust_decimated_plots/errors/spline_dec_50_train_faust_errors_80} &
	\includegraphics[height=120pt]{faust_decimated_plots/errors/feast_dec_50_train_faust_errors_80} &
	\includegraphics[height=120pt]{faust_decimated_plots/errors/feast_dual_dec_50_train_faust_errors_80} &
	\includegraphics[height=120pt]{faust_decimated_plots/errors/dual_dec_50_train_faust_xyz_errors_80}
	\\
	 & \small\textbf{Faust-Synthetic} & \small\textbf{MoNet} & \small\textbf{SplineCNN} & \small\textbf{FeaStNet} & \small{\textbf{FeaStNet--Dual}} & \small{\textbf{DualConvMax}}
	\\
	 & \small\textbf{Reference} & \small{(SHOT)} & \small{(XYZ)} & \small{(XYZ)} & \small{(XYZ)} & \small{(XYZ)}
	\\
	\end{tabular}
\caption{Visualizations of texture transfer and geodesic correspondence errors for a full resolution Faust-Synthetic test mesh (top two rows), and the same mesh decimated by 50\% (bottom two rows) of the Faust-Decimated dataset. Models are trained on the full resolution Faust-Synthetic meshes.
}
\label{fig:faust_decimated_errors}
\end{figure*}

\mypar{Transfer from Faust-Synthetic to decimated meshes}
In \fig{faust_decimated_errors}, we provide visualizations comparing the results of different methods trained on full resolution Faust-Synthetic meshes and tested on a full resolution test mesh and its $50\%$ decimated version. 
This figure complements Figure 7 of the main paper.

\begin{figure*}
	\resizebox{\textwidth}{!}{
	\begin{tabular}{cccccc}
	\includegraphics[height=100pt]{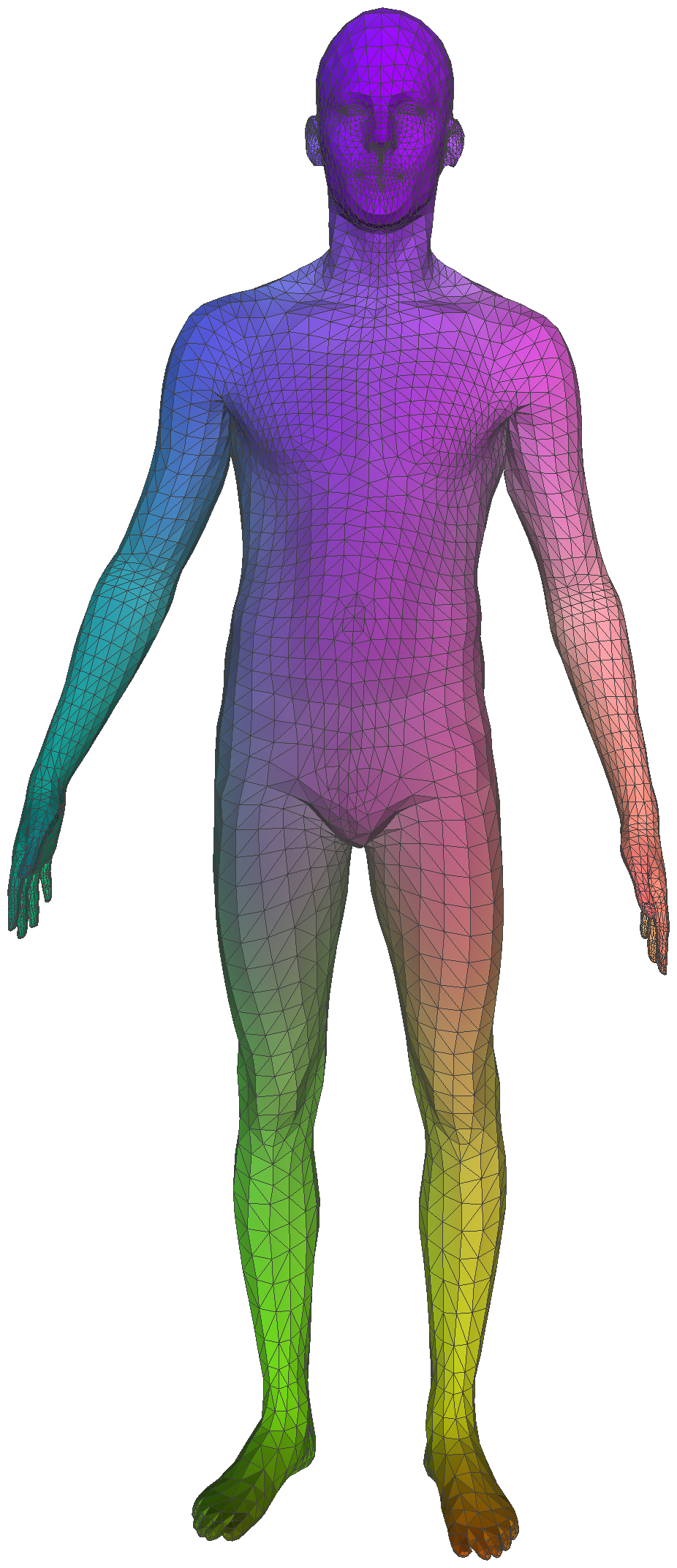} &
	\includegraphics[height=100pt]{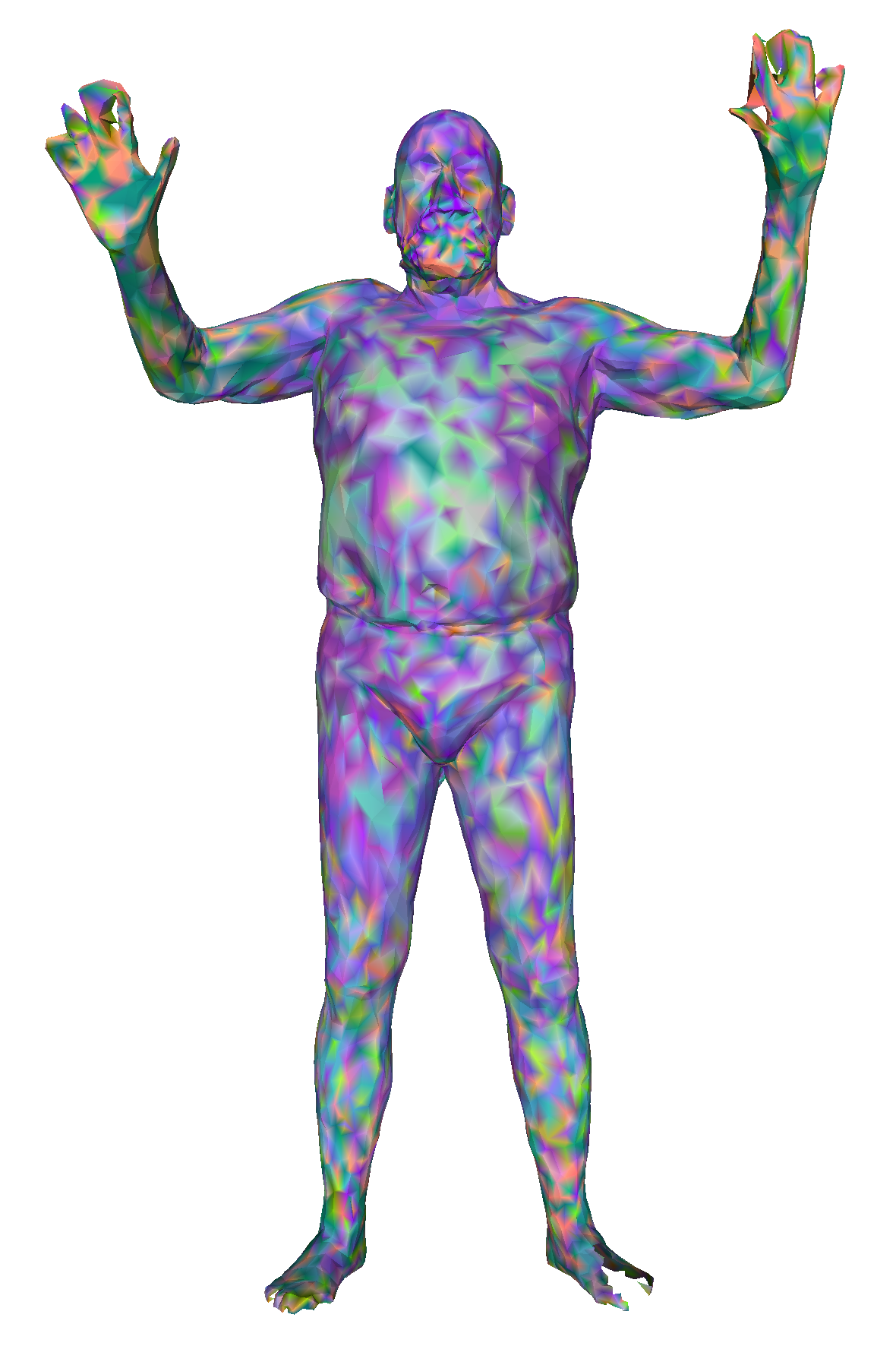} &
	\includegraphics[height=100pt]{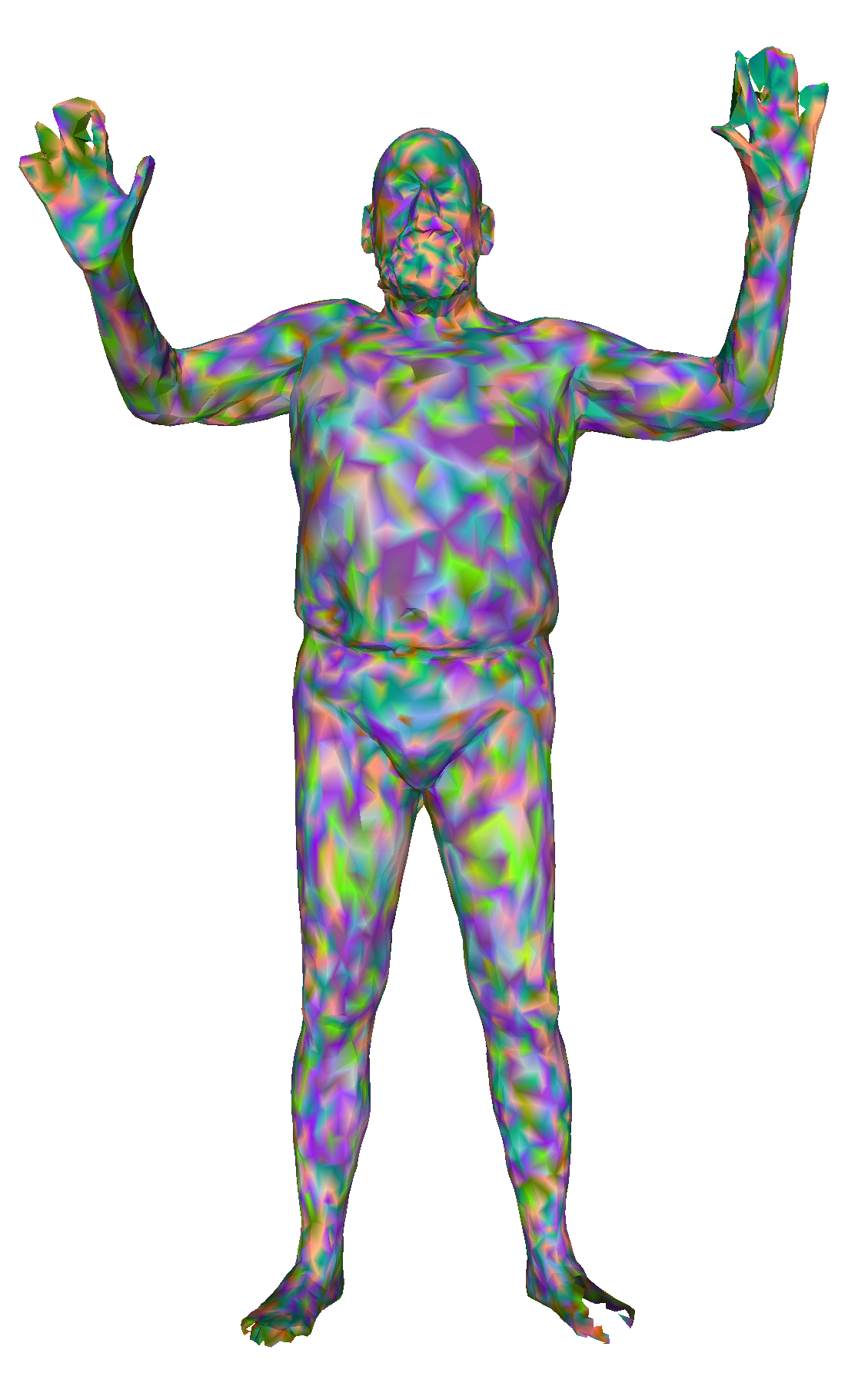} &
	\includegraphics[height=100pt]{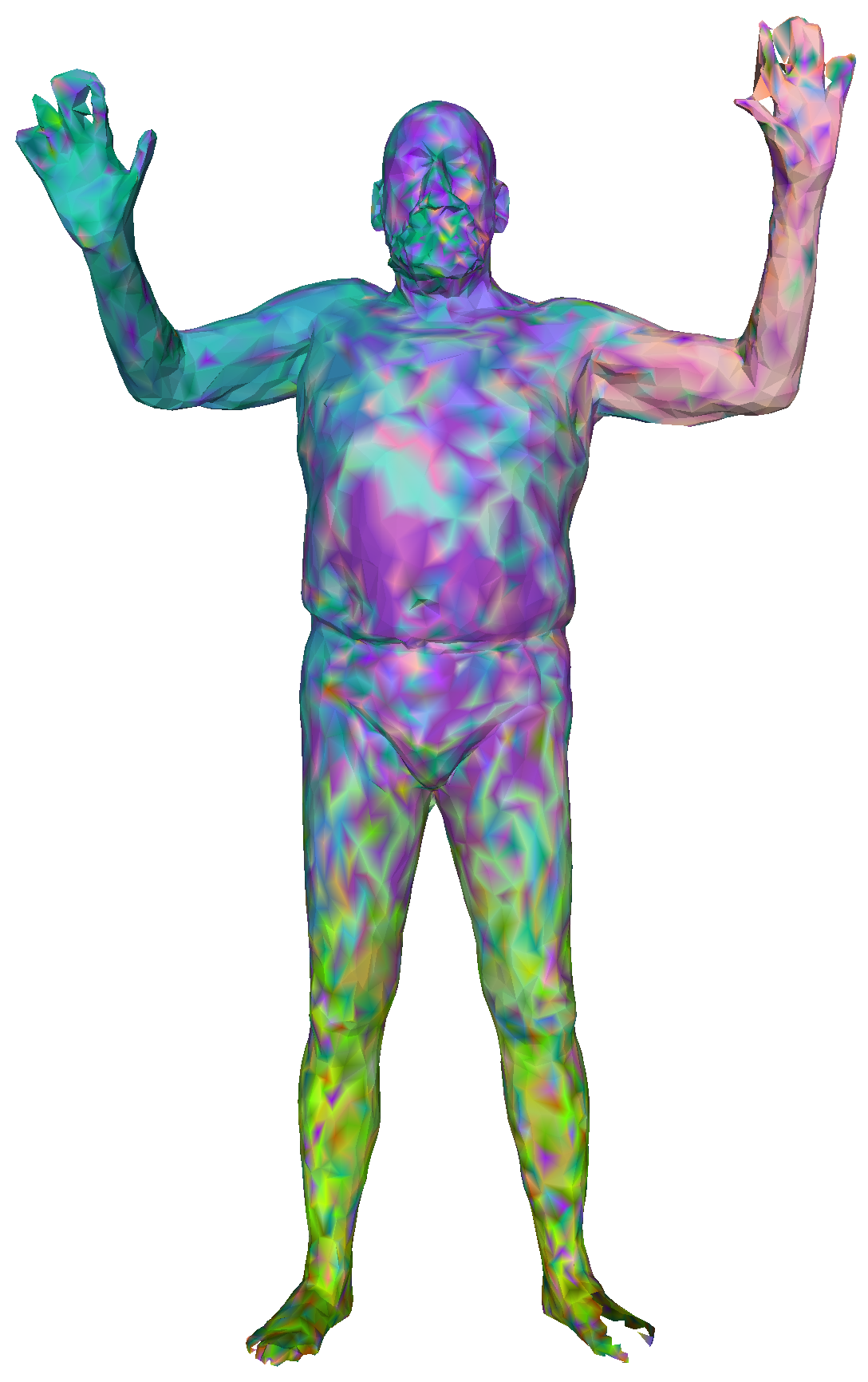} &
	\includegraphics[height=100pt]{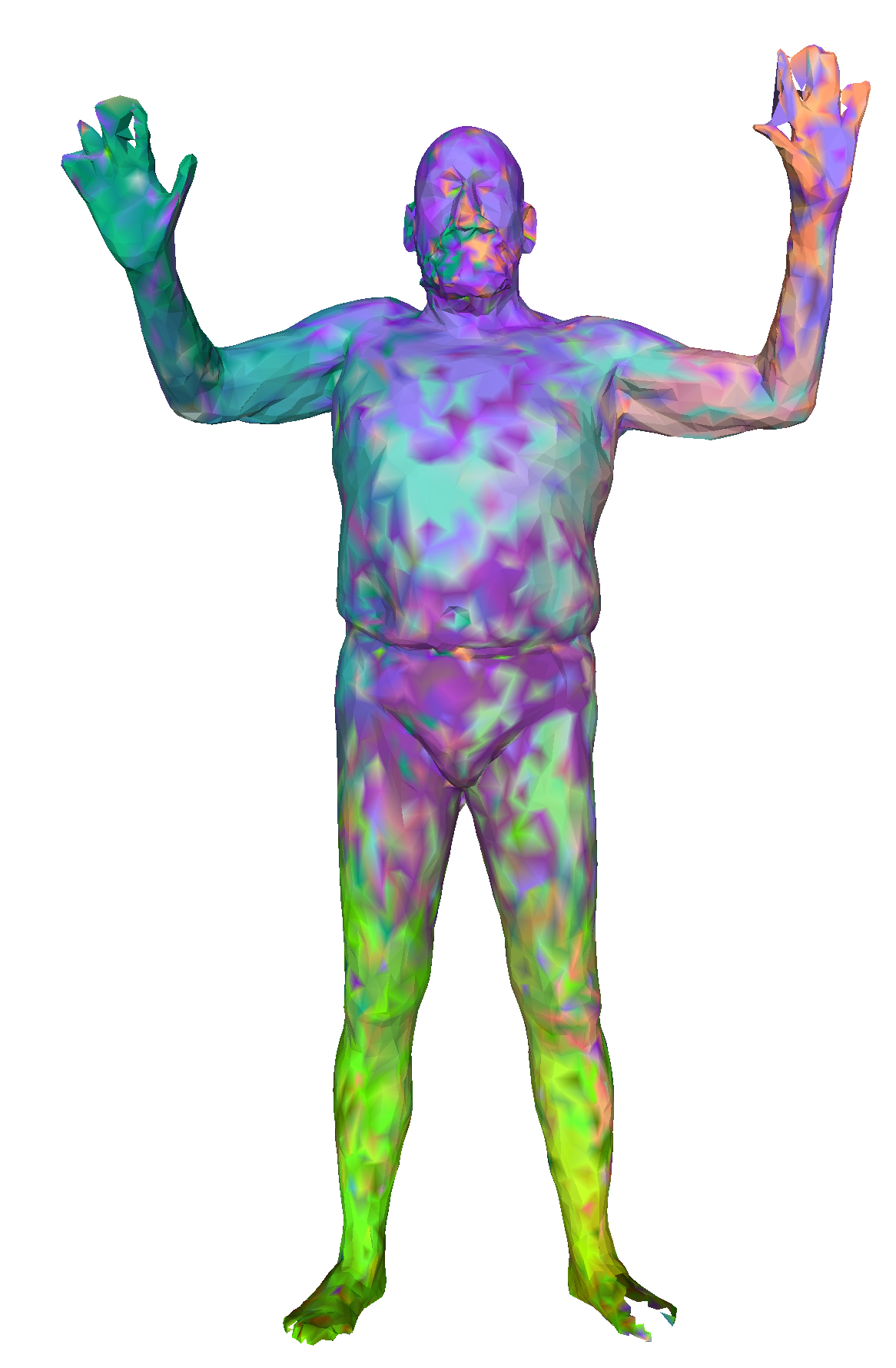} &
	\includegraphics[height=100pt]{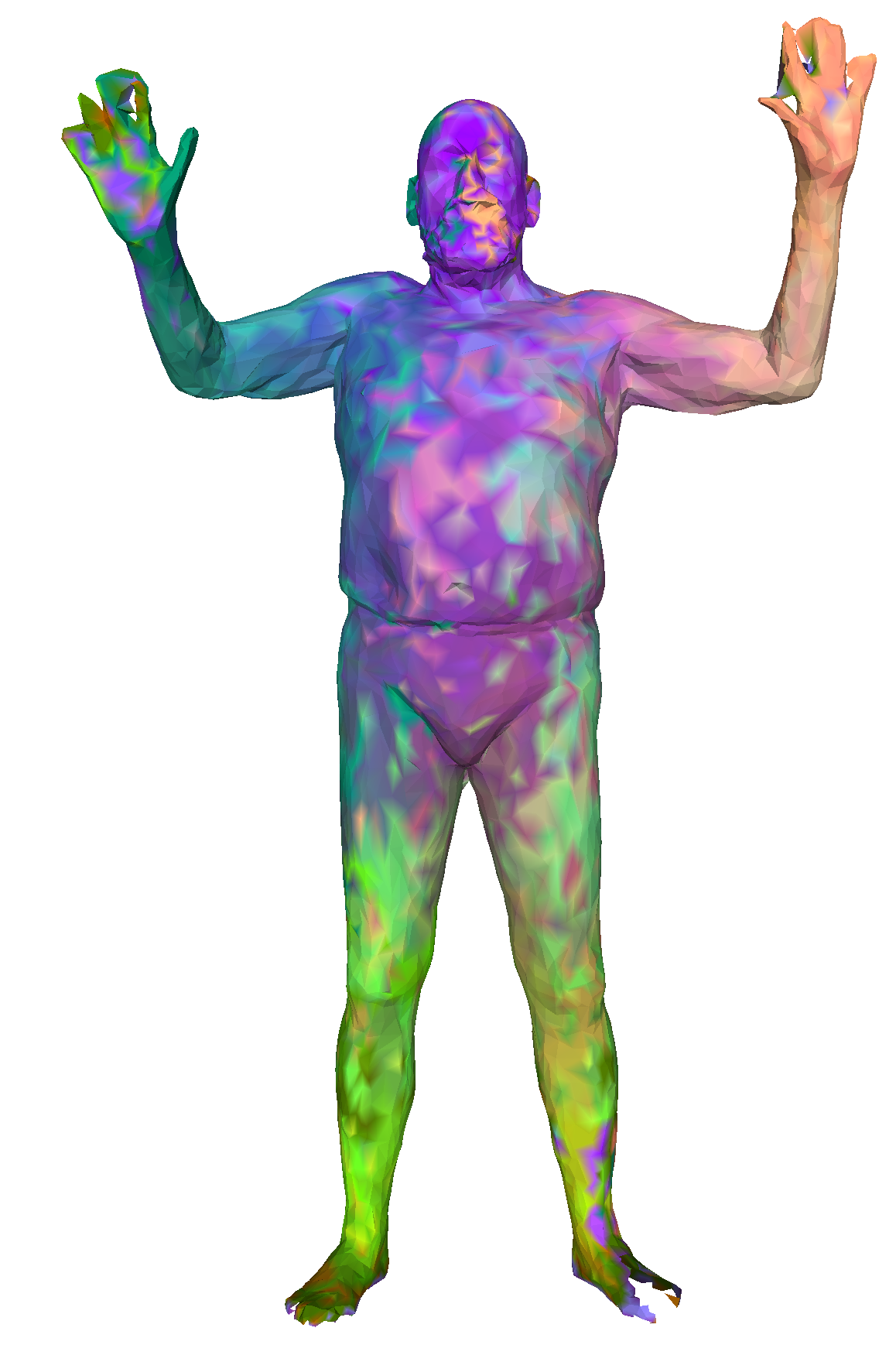}  
	\\
	\footnotesize\textbf{Faust-Synthetic} & \footnotesize\textbf{MoNet} & \footnotesize\textbf{SplineCNN} & \footnotesize\textbf{FeaStNet} & \footnotesize\textbf{FeaStNet--Dual} & \footnotesize\textbf{DualConvMax}
	\\
	\footnotesize\textbf{Reference} & \footnotesize{(SHOT)} & \footnotesize{(XYZ)} & \footnotesize{(XYZ)} & \footnotesize{(XYZ)} & \footnotesize{(XYZ)}
	\\
	\end{tabular}
}
\caption{Visualizations of texture transfer for Faust-Scan test results.
All methods have been trained on the Faust-Synthetic dataset.}
\label{fig:test_scan_train_faust}
\label{fig:test_remeshed_train_faust}
\end{figure*}

\mypar{Qualitative results on Faust-Scan}
We compare the results for methods trained on Faust-Synthetic and tested on Faust-Scan in \fig{test_scan_train_faust}.
This figure complements Figure 8 of the main paper.

\begin{figure*}
	\resizebox{\textwidth}{!}{
	\begin{tabular}{c|ccc|cc}
	\textbf{Faust-Remeshed} & \textbf{MoNet} & \textbf{SplineCNN} & \textbf{FeaStNet} & \textbf{FeaStNet--Dual} & \textbf{DualConvMax}
	\\
	\textbf{Ref} & \small{(SHOT)} & \small{(XYZ)} & \small{(XYZ)} & \small{(XYZ+Normal)} & \small{(XYZ+Normal)}
	\\
	\includegraphics[scale=0.053]{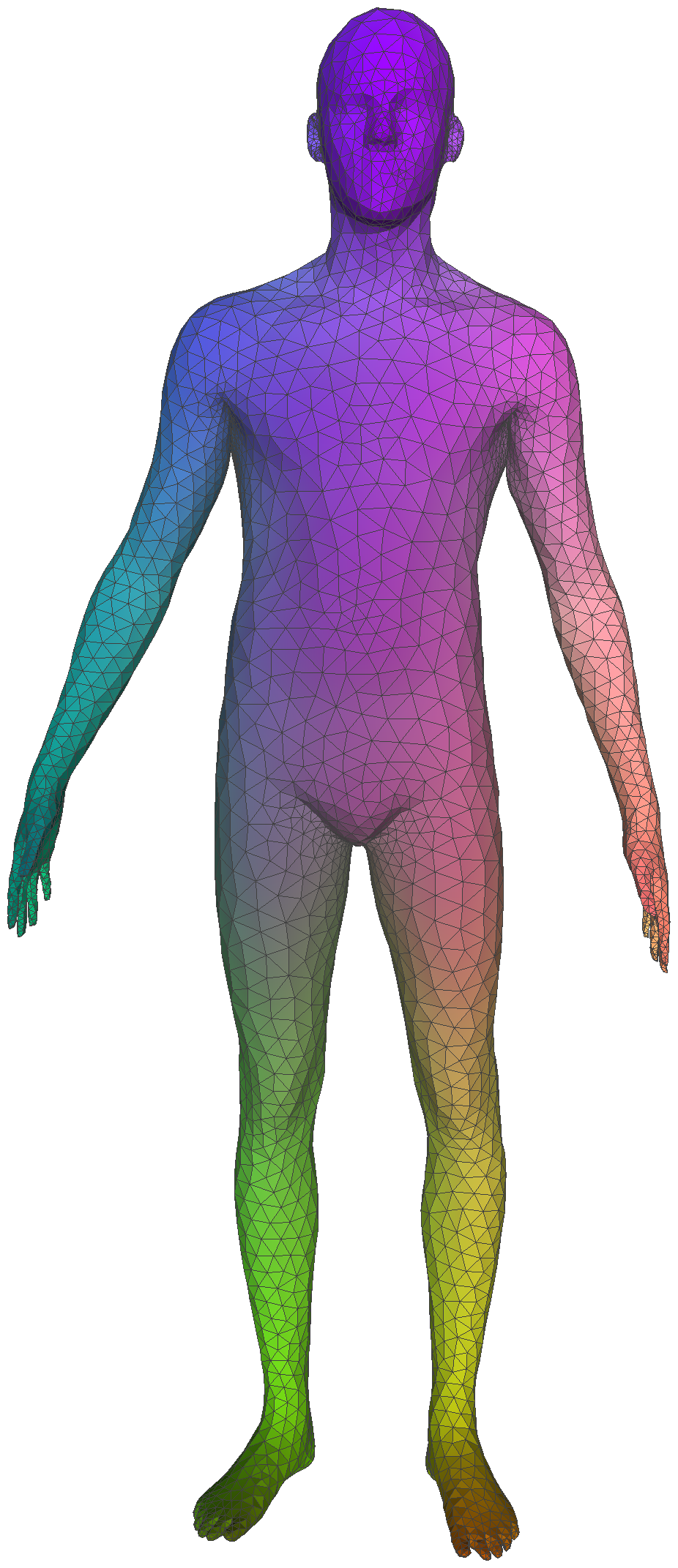} &
	\includegraphics[scale=0.08]{remeshed_faust_plots/83_monet_train_remeshed} &
	\includegraphics[scale=0.08]{remeshed_faust_plots/83_spline_train_remeshed} &
	\includegraphics[scale=0.08]{remeshed_faust_plots/83_feast_train_remeshed} &
	\includegraphics[scale=0.08]{remeshed_faust_plots/83_feast_dual_xyz_normal_train_remeshed} &
	\includegraphics[scale=0.08]{remeshed_faust_plots/83_dual_max_xyz_normal_train_remeshed}
	\\
	\includegraphics[height=60px]{figures/errors_colormap} &
	\includegraphics[scale=0.08]{remeshed_faust_plots/errors/83_monet_train_remeshed} &
	\includegraphics[scale=0.08]{remeshed_faust_plots/errors/83_spline_train_remeshed} &
	\includegraphics[scale=0.08]{remeshed_faust_plots/errors/83_feast_train_remeshed} &
	\includegraphics[scale=0.08]{remeshed_faust_plots/errors/83_feast_dual_xyz_normal_train_remeshed} &
	\includegraphics[scale=0.08]{remeshed_faust_plots/errors/83_dual_max_xyz_normal_train_remeshed}
	\\
	\includegraphics[scale=0.053]{remeshed_faust_plots/remeshed_faust_ref} &
	\includegraphics[scale=0.08]{remeshed_faust_plots/97_monet_train_remeshed} &
	\includegraphics[scale=0.08]{remeshed_faust_plots/97_spline_train_remeshed} &
	\includegraphics[scale=0.08]{remeshed_faust_plots/97_feast_train_remeshed} &
	\includegraphics[scale=0.08]{remeshed_faust_plots/97_feast_dual_xyz_normal_train_remeshed} &
	\includegraphics[scale=0.08]{remeshed_faust_plots/97_dual_max_xyz_normal_train_remeshed} 
	\\
	\includegraphics[height=60px]{figures/errors_colormap} &
	\includegraphics[scale=0.08]{remeshed_faust_plots/errors/97_monet_train_remeshed} &
	\includegraphics[scale=0.08]{remeshed_faust_plots/errors/97_spline_train_remeshed} &
	\includegraphics[scale=0.08]{remeshed_faust_plots/errors/97_feast_train_remeshed} &
	\includegraphics[scale=0.08]{remeshed_faust_plots/errors/97_feast_dual_xyz_normal_train_remeshed} &
	\includegraphics[scale=0.08]{remeshed_faust_plots/errors/97_dual_max_xyz_normal_train_remeshed}
	\\
	\textbf{Faust-Remeshed} & \textbf{MoNet} & \textbf{SplineCNN} & \textbf{FeaStNet} & \textbf{FeaStNet--Dual} & \textbf{DualConvMax}
	\\
	\textbf{Ref} & \small{(SHOT)} & \small{(XYZ)} & \small{(XYZ)} & \small{(XYZ+Normal)} & \small{(XYZ+Normal)} \\
	\end{tabular}
}
\caption{Visualizations of texture transfer for Faust-Remeshed test results trained on Faust-Remeshed dataset using the state-of-the-art methods and our method. The vertices for which the ground-truth is missing are colored blue in the error meshes.
}
\label{fig:test_remeshed_faust_train_remeshed}
\end{figure*}

\mypar{Qualitative results on Faust-Remeshed}
In \fig{test_remeshed_faust_train_remeshed}, we visualize the texture transfer from the Faust-Remeshed reference shape to Faust-Remeshed test shapes and correspondence errors for these methods. 
We represent the vertices for which the ground-truth is missing as blue in the  display.
This figure corresponds to the evaluation in Table 3 of the main paper.

\begin{figure*}
    \centering
	\begin{tabular}{cc|cc}
	\includegraphics[height=120pt]{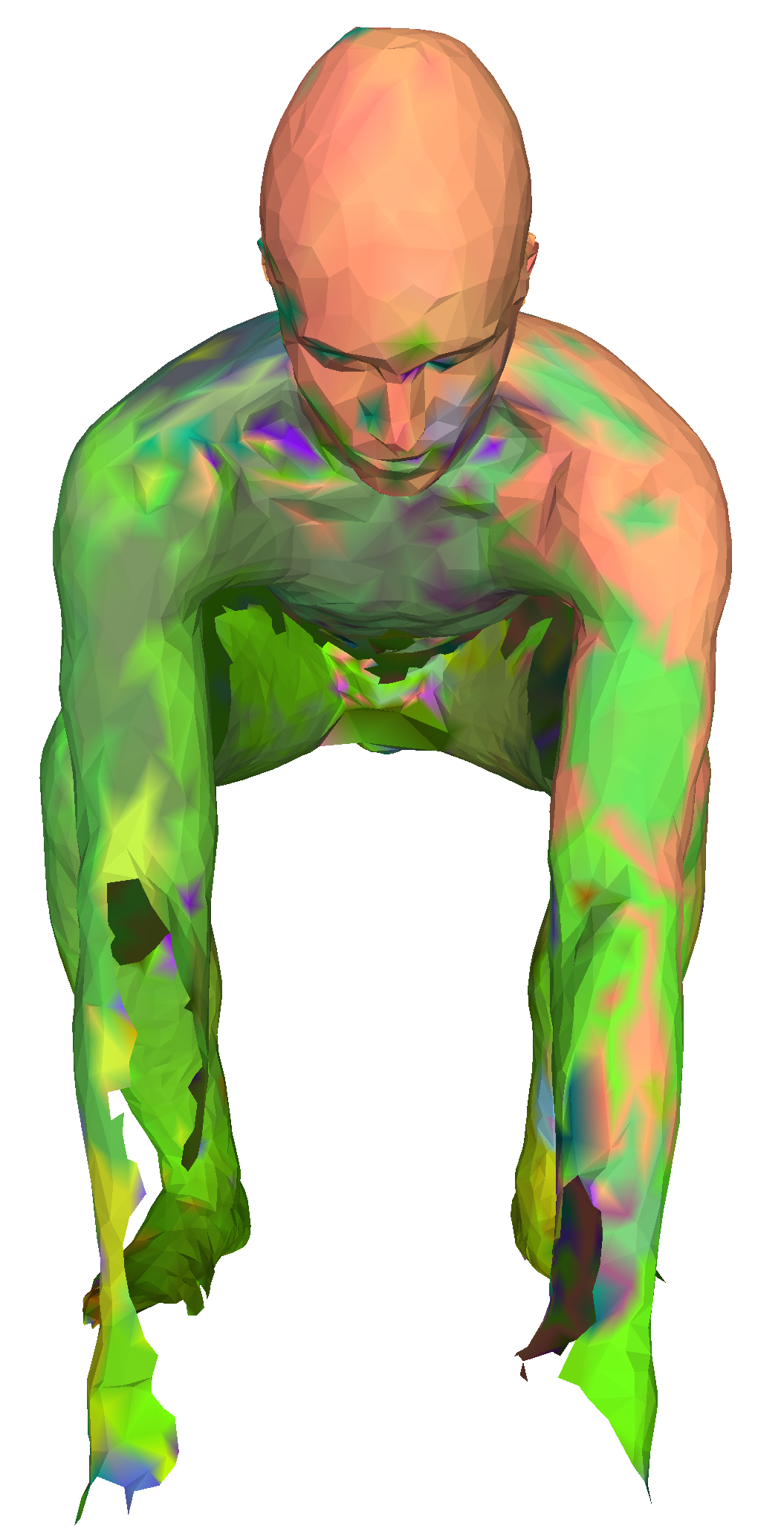} &
	\includegraphics[height=120pt]{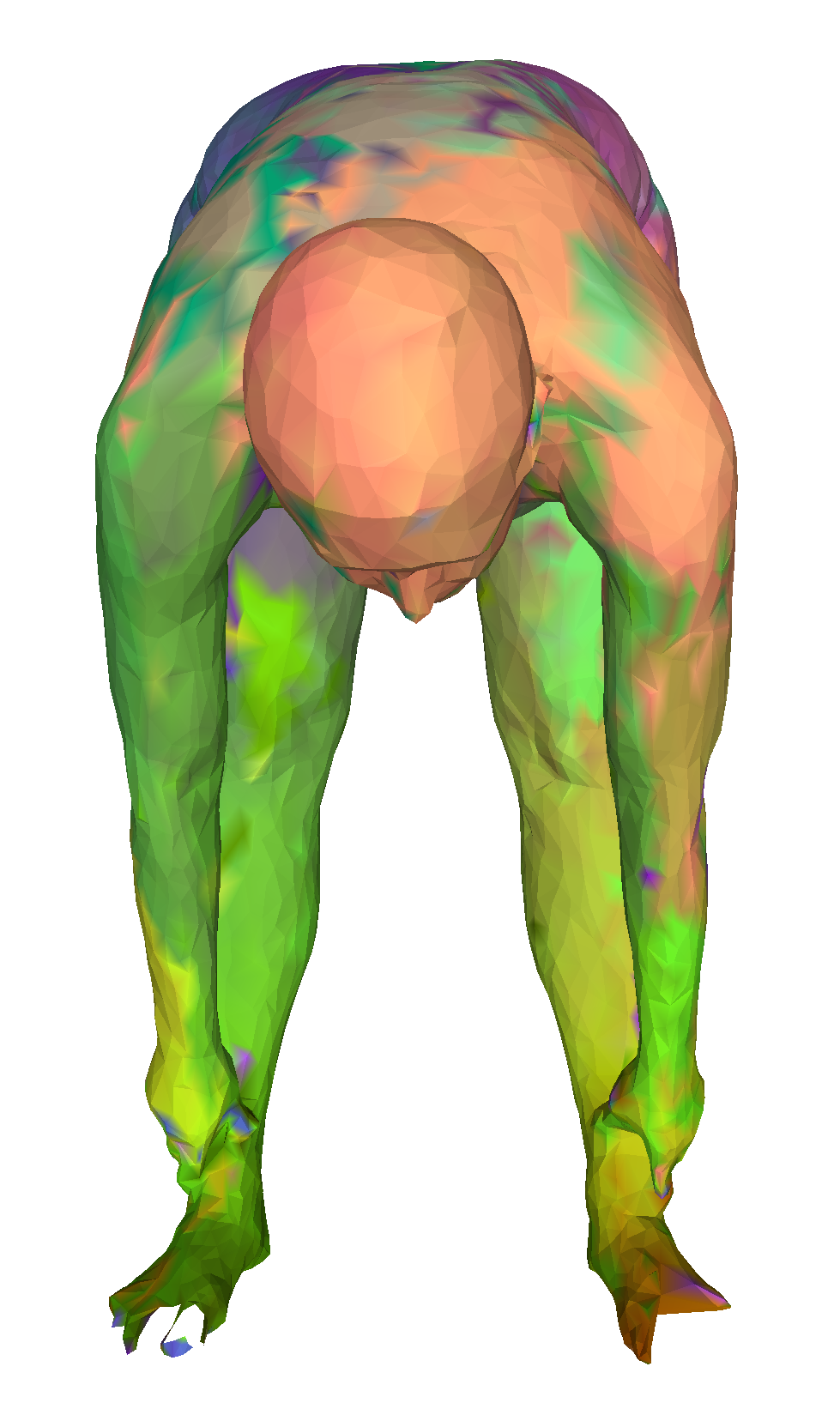} &
	\includegraphics[height=120pt]{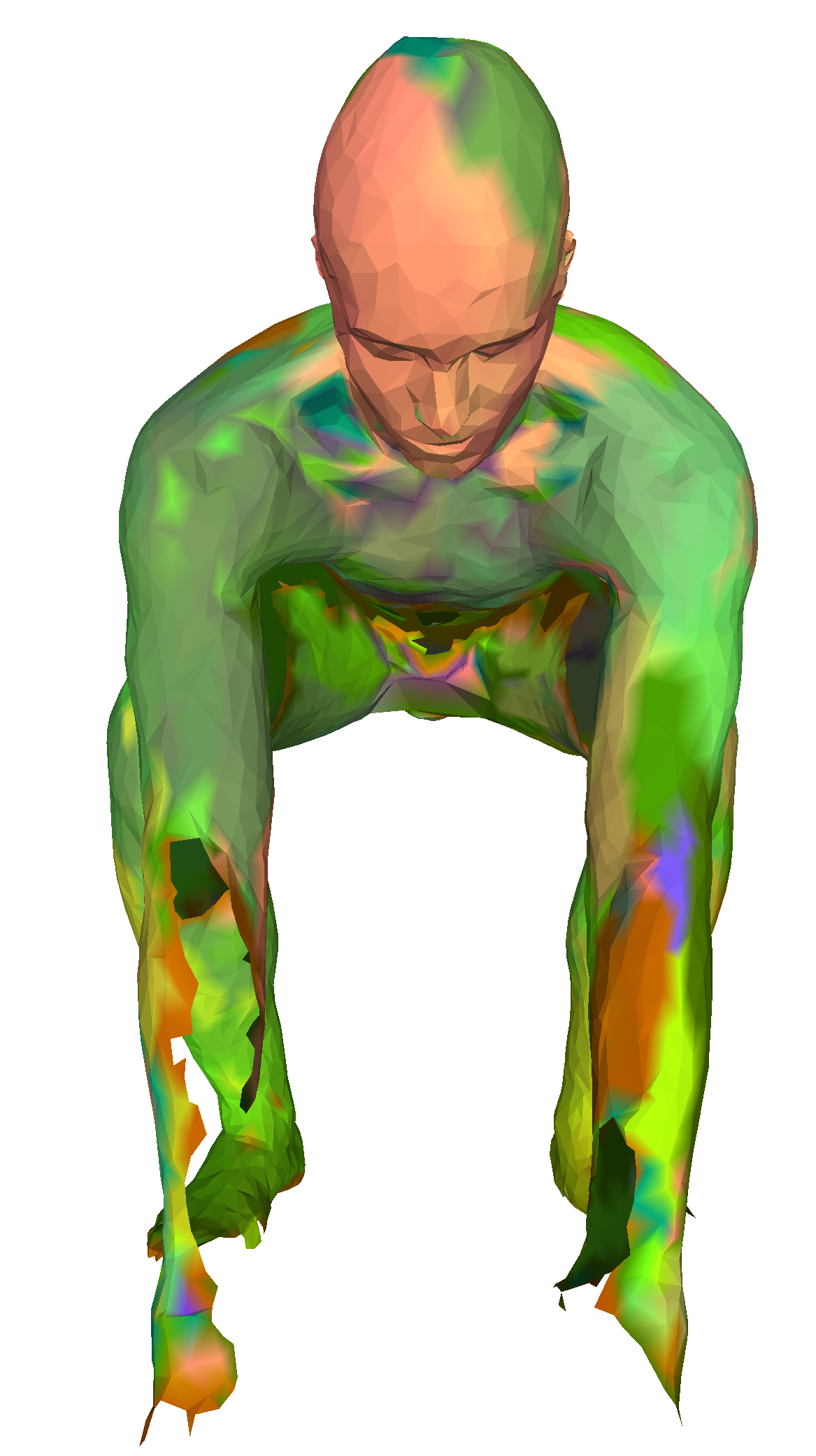} &
	\includegraphics[height=120pt]{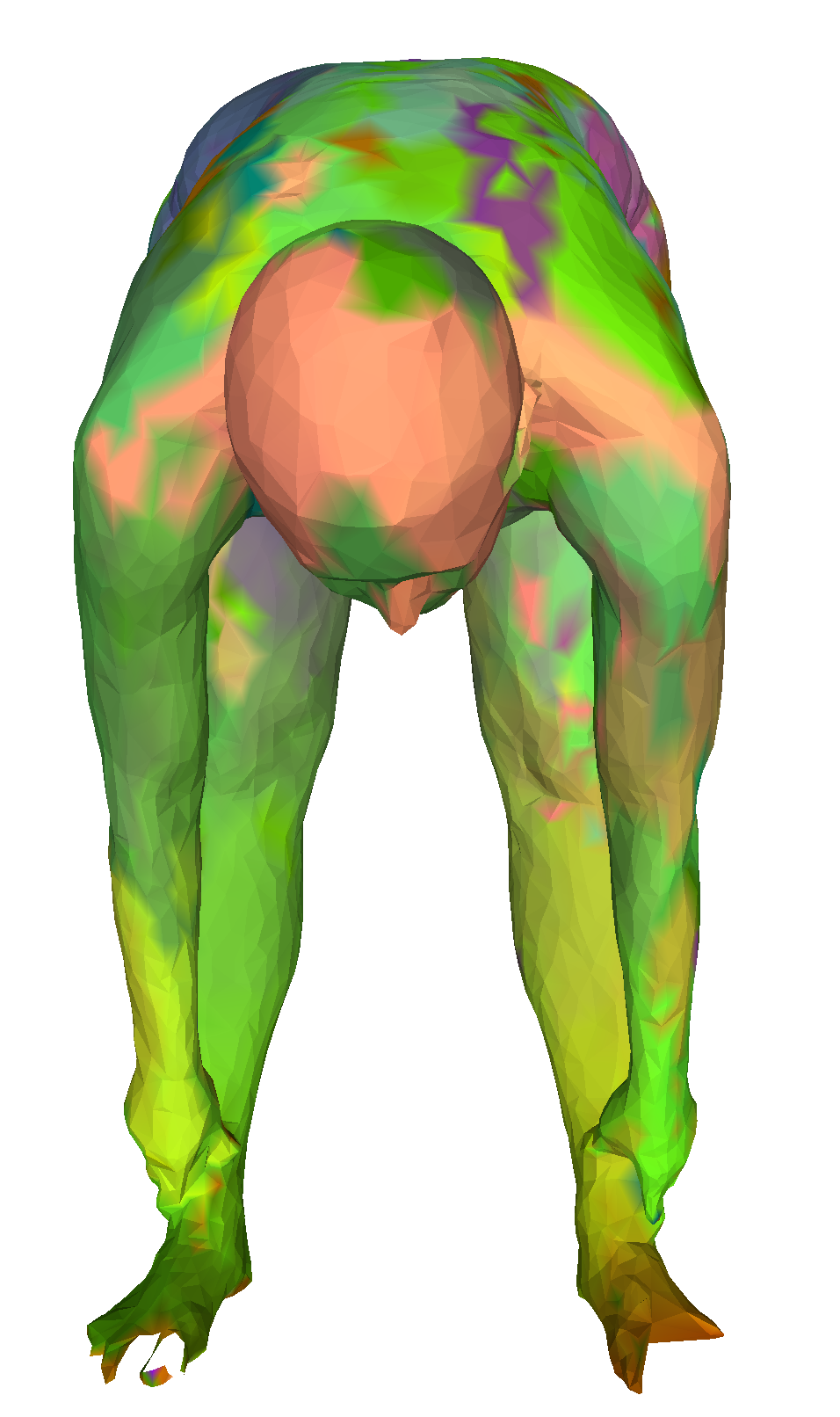}
	\\
	\multicolumn{2}{c|}{\small\textbf{FeaStNet--Dual}} & \multicolumn{2}{c}{\small\textbf{DualConvMax}}
	\end{tabular}
\caption{Visualization of failure cases on Faust-Scan of models trained on the Faust-Remeshed dataset using XYZ+Normal as input.
}
\label{fig:limitations}
\end{figure*}

\mypar{Limitations}
Failures  of the dual-based models tend to appear on test poses in Faust-Scan that are very different from the, mostly upright, training poses in Faust-Remeshed.
We show two such instances in \fig{limitations}. 
These bent poses is particularly problematic for the correspondences on the upper body.
This figure complements the results shown in figures 2 and 9  of the main paper.

{\small
\bibliographystyle{ieee_fullname}
\bibliography{jjv,more}
}

%% file: 3dvheader.tex
\usepackage{3dv}
\usepackage{times}
\usepackage{epsfig}
\usepackage{graphicx}
\usepackage{amsmath}
\usepackage{amssymb}

\usepackage{booktabs}
\usepackage{xcolor}
\usepackage{multirow}
\input{macros}

\newcommand{\jkb}[1]{{{#1}}} 
\newcommand{\nitika}[1]{{{#1}}}

\def\mypar#1{{\noindent\bf #1.}\hspace{1mm}}
\def\ssect#1{Section~\ref{ssec:#1}}

\usepackage[pagebackref=true,breaklinks=true,colorlinks,bookmarks=false]{hyperref}

\threedvfinalcopy 

\def\threedvPaperID{135} 

\ifthreedvfinal\fi

\author{
Nitika Verma$^{1}$ \quad\quad
Adnane Boukhayma$^{2}$ \quad\quad
Edmond Boyer$^{1}$ \quad\quad
Jakob Verbeek$^{3}$ 
\vspace{1mm}
\\
 $^{1}$Inria, Univ. Grenoble Alpes, CNRS, Grenoble INP, LJK, France\\ 
 $^{2}$Inria, Univ. Rennes, CNRS, IRISA, M2S, France\\
 $^{3}$Facebook AI Research   
}


%% file: macros.tex
\newif\ifrevfinal
\revfinalfalse
\def\rev[#1][#2]{\ifrevfinal #2 \else {\color{blue} \sout{#1}} {\bf \color{red} #2} \fi}
\usepackage{color}
\usepackage[normalem]{ulem}
\usepackage{amsmath,amssymb,amsbsy,xspace}

\def\R{{\rm I\!R}}                            	

\usepackage{amsmath}

\def\<{\langle}
\def\>{\rangle}

\def\be{\begin{equation}}
\def\ee{\end{equation}}
\def\bea{\begin{eqnarray}}
\def\eea{\end{eqnarray}}
\def\fig#1{Figure~\ref{fig:#1}}
\def\tab#1{Table~\ref{tab:#1}}
\def\sect#1{Section~\ref{sec:#1}}

\makeatletter
\gdef\SetFigFont#1#2#3{%
  \reset@font\fontsize{10}{12pt}%
  \selectfont%
}
\let\eqnarray@=\eqnarray \let\endeqnarray@=\endeqnarray
\def\eqnarray{\bgroup\arraycolsep=2pt\eqnarray@}
\def\endeqnarray{\endeqnarray@\egroup}
\makeatother

\def\em{\slshape}

\makeatletter
\DeclareRobustCommand\onedot{\futurelet\@let@token\@onedot}
\def\@onedot{\ifx\@let@token.\else.\null\fi\xspace}

\def\eg{\emph{e.g}\onedot} 
\def\ie{\emph{i.e}\onedot} 
\def\cf{\emph{c.f}\onedot} 
 \def\vs{\emph{vs}\onedot} 
\def\wrt{w.r.t\onedot}

\newcounter{iictr}
\def\@iia[#1]{\setcounter{iictr}{#1}\@iib}
\def\@iib{\iii[\roman{iictr}]\xspace}
\def\ii{\@ifnextchar[{\@iia}{\stepcounter{iictr}\@iib}}
\def\iii[#1]{({\em#1\/})}
\makeatother


%% file: sec_abstract.tex
\begin{abstract}

Convolutional networks have been extremely successful for regular data structures such as  2D images and 3D voxel grids.
The transposition to meshes is, however, not straightforward due to their irregular structure.
We explore how the dual, face-based representation of triangular meshes can be leveraged as a data structure for graph convolutional networks.
In the dual mesh, each node (face) has a fixed number of neighbors, which makes the networks less susceptible to overfitting on the mesh topology, and also allows the use of input features that are naturally defined over faces, such as surface normals and face areas.
We evaluate the dual approach on the shape correspondence task on the Faust human shape dataset and variants of it with different mesh topologies.
Our experiments show that results of graph convolutional networks improve when defined over the dual rather than primal mesh. 
Moreover, our models that explicitly leverage the neighborhood regularity of dual meshes allow improving results further while being more robust to changes in the mesh topology.

\end{abstract}

%% file: sec_intro.tex
\section{Introduction}

The success of  convolutional neural networks for recognition  in 2D images \cite{he16eccv,krizhevsky12nips,simonyan15iclr} has spurred efforts to transfer these results to the analysis of 3D shape data.
One of the most direct approaches is to extend the 2D convolutions to 3D voxel grids~\cite{choy16eccv,maturana15iros,tatarchenko17iccv}. 
Voxel grids are, however, inefficient in that they are extrinsic and quantize space rather than the shape itself.
While intrinsic representations such as point clouds and meshes are more attractive to model shapes since they directly approximate the shape itself, the formulation of deep neural networks on such irregular data structures is more complex.
Point clouds provide a simple orderless data structure, and neural networks can be constructed by combining local per-point operations with global permutation invariant operations~\cite{klokov17iccv,qi17cvpr}.
In our work, we focus on 3D mesh representations, which  offer a topological graph structure on top of the vertex positions, allowing for a compact and accurate surface characterization.


A variety of approaches have been explored in previous work to define deep neural networks on irregularly structured meshes, where the number of neighbors can change from one vertex to another. 
Most of these methods treat meshes as graphs, where the nodes of the graph are the mesh vertices connected by the edges of the surface triangles. 
To process data on such graphs,  they apply global spectral operators~\cite{bruna14iclr,chen2018fastgcn, defferrard16nips,huang2018adaptive,kipf17iclr, levie2018cayleynets} or local spatial filters~\cite{fey18cvpr,monti17cvpr,verma18cvpr}. Other methods  are  formulated by taking into account properties specific to meshes, such as \cite{gong2019spiralnet++,lim2018simple,milano2020primal}.
We discuss related work in more detail in \sect{related}.


\begin{figure}
    \resizebox{\columnwidth}{!}{
        \includegraphics[width=70pt]{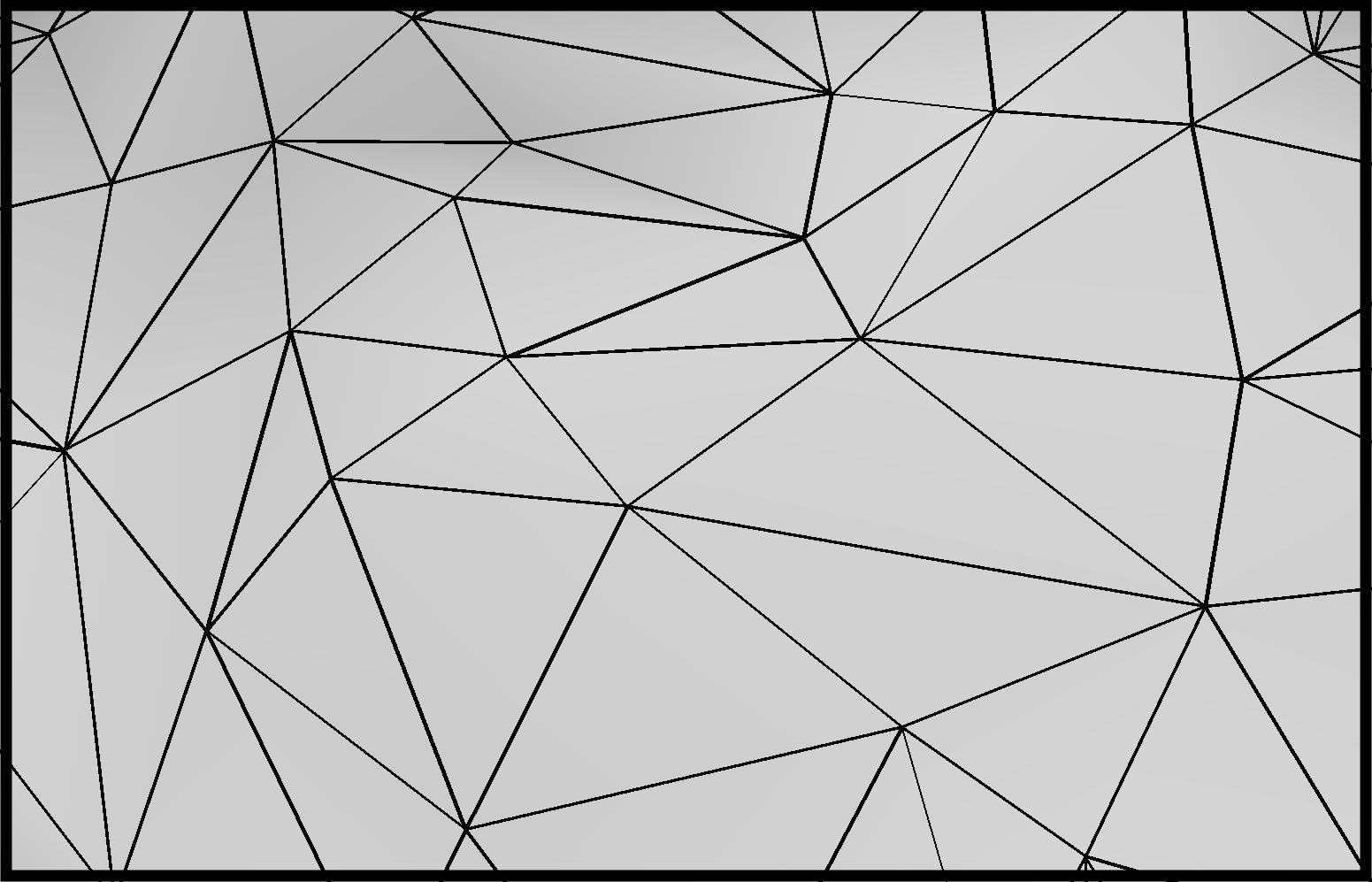}
        \includegraphics[width=70pt]{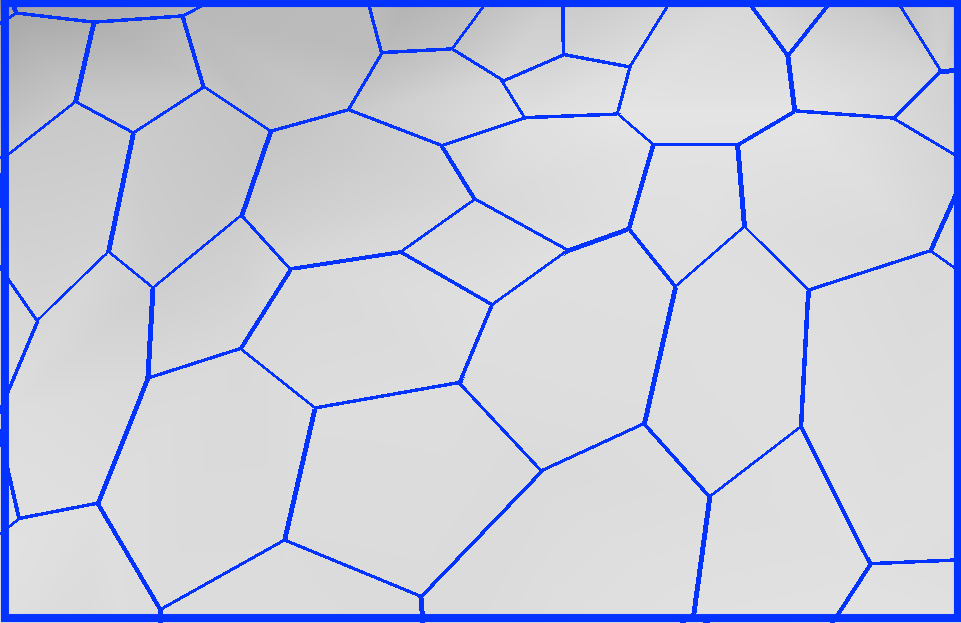}
    }
\caption{Illustration of a triangular primal mesh (left) and its dual  (right). 
Note that every vertex in the dual has exactly three neighbors, while the number of neighbors is not constant in the primal.
}
\label{fig:orig_dual_faust}
\end{figure}

We study the use of the dual mesh defined over the faces, where each vertex represents a face and is connected to the incident faces, see \fig{orig_dual_faust}. 
Using the faces rather than the vertices to represent the data, it is natural to use input features such as the face normal, in combination with the face center location. 
Moreover, for watertight triangular meshes, each vertex has exactly three neighbors in the dual mesh, which we exploit to define a  convolution operator called DualConvMax on the dual mesh.

\begin{figure}
	\resizebox{\columnwidth}{!}{
	\begin{tabular}{cccc}
	\includegraphics[height=200pt]{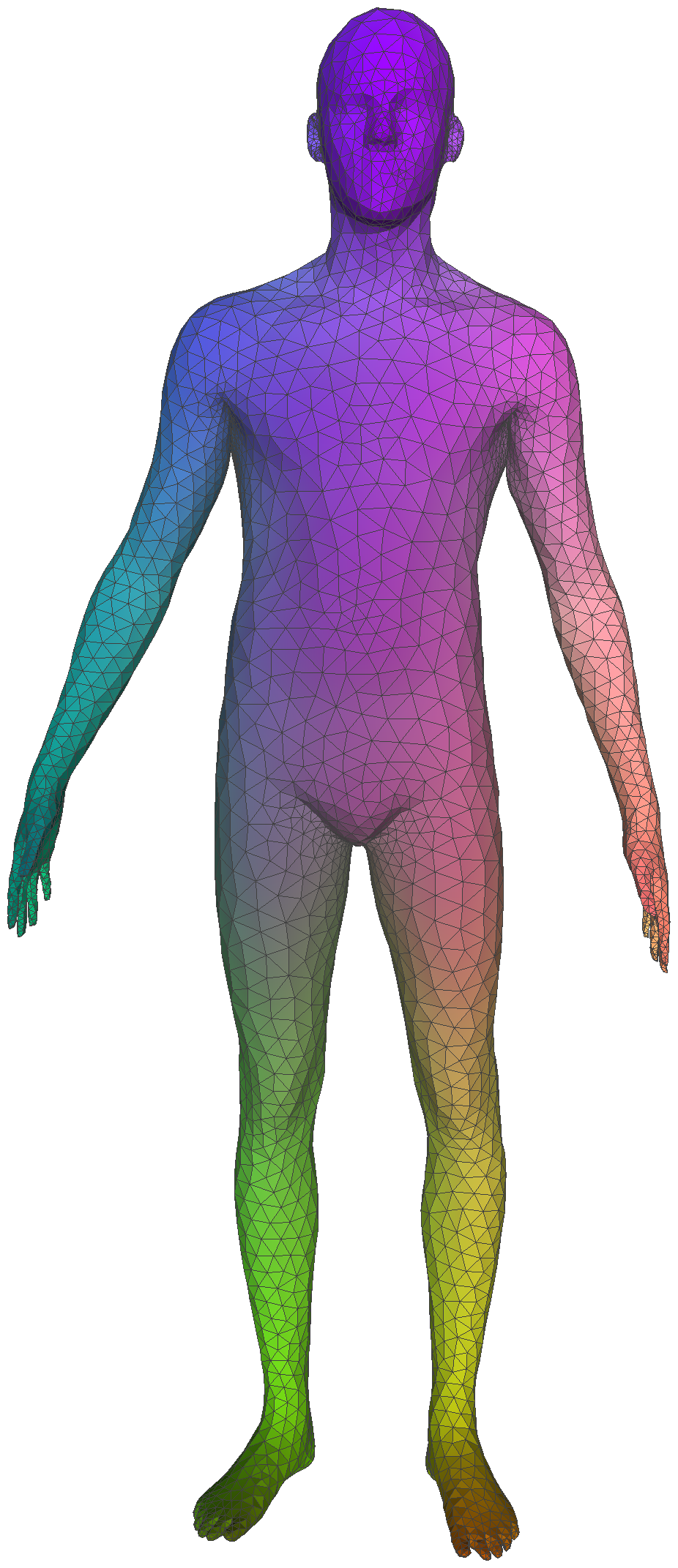} &
	\includegraphics[height=200pt]{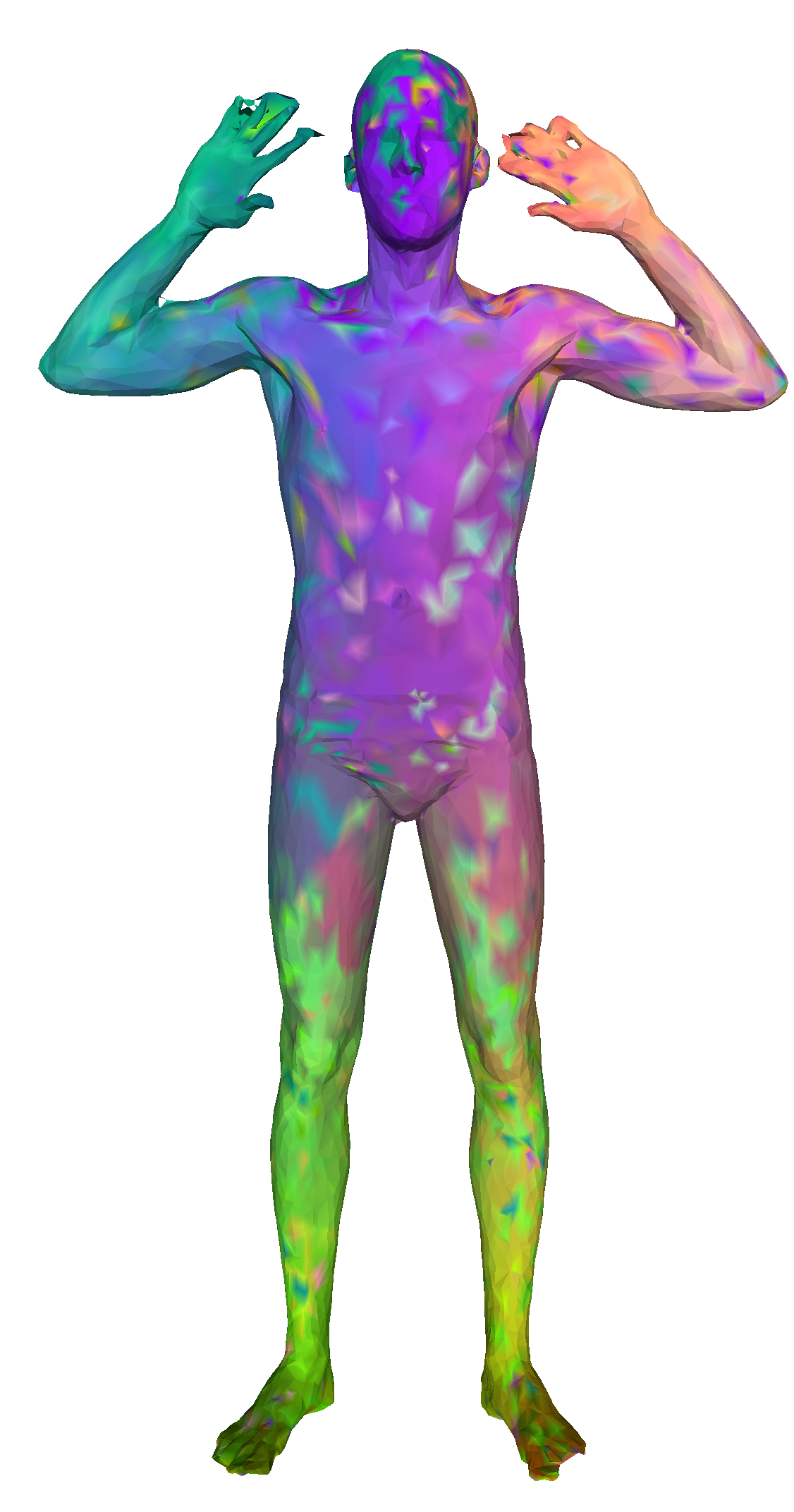} &
	\includegraphics[height=200pt]{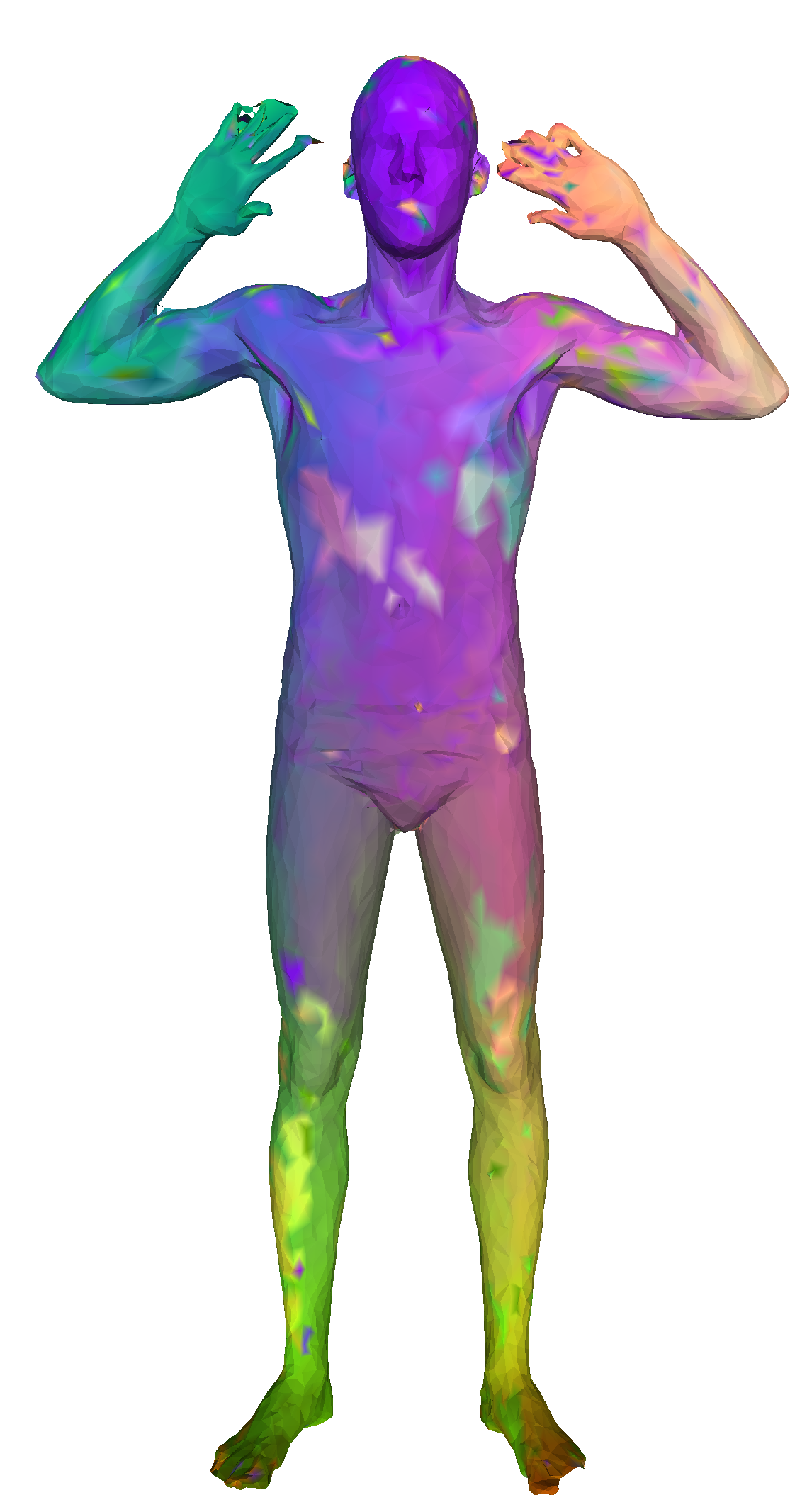} &
	\includegraphics[height=200pt]{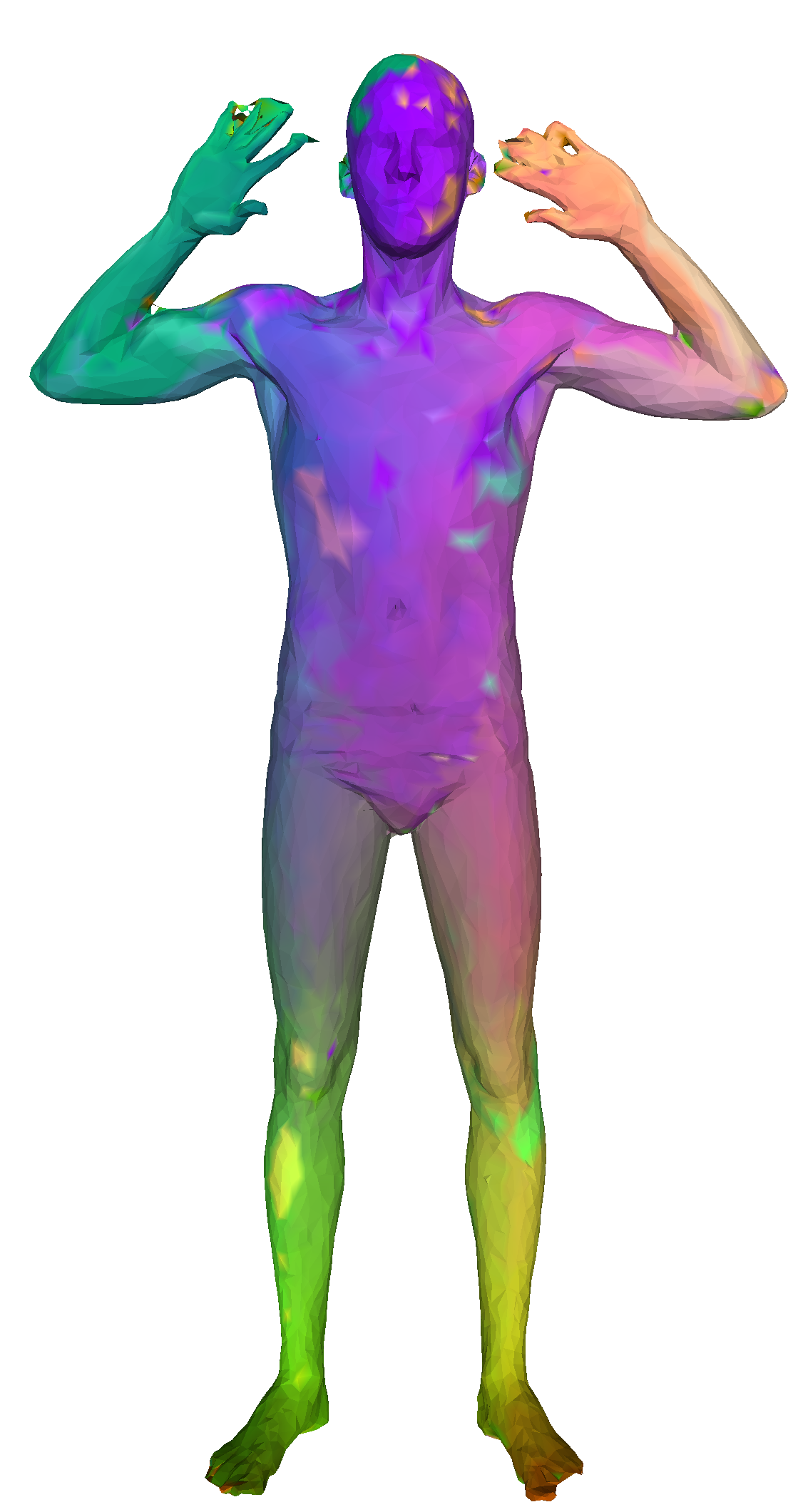} 
	\\
	\LARGE\textbf{Reference} & \LARGE\textbf{FeaStNet} & \LARGE\textbf{FeaStNet--Dual} & \LARGE\textbf{DualConvMax}
	\\
	\end{tabular}
}
\caption{Visualizations of texture transfer from a reference shape to decimated raw Faust scans using primal mesh based method FeaStNet, its dual variant FeaStNet--Dual and our proposed DualConvMax. 
All models were trained on the Faust-Remeshed data.
}
\label{fig:teaser}
\end{figure}


We conduct shape correspondence experiments on the Faust human shape dataset~\cite{bogo14cvpr}.
However, the meshes in the Faust dataset all share the same topology,
which is undesirable as it is not representative of real-world data where shapes have differently structured meshes. 
Therefore, we also consider more challenging evaluation setups for our approach and to compare to previous work. 
First, we consider evaluating models trained on Faust on variants of the meshes which have been decimated to different degrees.
Second, we train and test the models on a re-meshed version of Faust~\cite{ren2018continuous}, in a setup where the mesh structure varies both in training and testing. In both cases, we also  test on decimated versions of the raw Faust scans.

We find that existing graph convolutional methods improve in the dual domain 
due to the addition of face-based features. 
Our DualConvMax model, which leverages the regularity of the dual mesh explicitly, 
further improves performance.
Moreover, we find that the dual-based models transfer considerably better in settings where the train and test data have different mesh topologies.

Qualitative correspondence results when applying the learned models to the original raw Faust scans  confirm the  quantitative results, see \fig{teaser}.   
In summary, our main contributions are the following:
\begin{itemize}
  \setlength\itemsep{0em}
    \item we propose  the DualConvMax layer to build convolutional networks over the face-based dual mesh;
    \item we propose a comparative evaluation of various input  features on the dual and their combinations;
    \item when train and test topologies differ, we find improved  performance using our dual approaches.
\end{itemize}


%% file: sec_related.tex
\section{Related work}
\label{sec:related}

We briefly review related work on deep learning for mesh data, based on spectral and spatial graph networks, as well as geometry-aware methods.  
We refer the reader to~\cite{wu2020comprehensive,zhang2019graph} for more extensive overviews of graph neural networks.

\mypar{Spectral methods}
Spectral graph convolutional networks are based on graph signal processing, for example, by extending convolutions to graphs using Laplacian Eigen-decomposition~\cite{bruna14iclr}.
In order to address the challenges posed by the high computational cost of this approach,  Chebyshev K-polynomials can be used to define localized and efficient convolution filters~\cite{defferrard16nips}. 
A simplified variant uses a first-order approximation of the Chebyshev expansion~\cite{kipf17iclr}. 
Following this seminal work, several other approaches have been proposed~\cite{chen2018fastgcn,huang2018adaptive,levie2018cayleynets}. However, spectral-based methods do not generalize well across domains with different graph structures. 
Consequently, they are primarily helpful in inferring node properties in situations where the graph during training and testing is the same \cite{bouritsas2019neural,ranjan2018generating,sen2008collective}, and less suitable for tasks where different graphs are considered during training and testing such as in  3D shape processing~\cite{hanocka19tog,ren2018continuous}. 

\mypar{Spatial methods}
Where spectral methods operate globally, spatial methods compute features by aggregating information locally, similar to traditional CNNs. However, this is not straightforward for mesh data due to their irregular local structures: (i) the number of neighbors per node may vary, and (ii) even if the number of neighbors is fixed, there might not be a consistent ordering among them. 
To alleviate these challenges, patch-operator based methods \cite{boscaini16nips,masci15iccvw} have been proposed where local patches are extracted using geodesic local polar coordinates and anisotropic heat kernels, respectively. 
Patch extraction has also been parameterized by mixtures of Gaussian kernels associated with local polar pseudo-coordinates~\cite{monti17cvpr}, using 
dynamically generated convolutional filter weights conditioned on edge attributes neighboring the vertices~\cite{Simonovsky2017}, or with
convolutional filters  based on B-spline bases with  learnt control values~\cite{fey18cvpr}. 
FeaStNet~\cite{verma18cvpr}  learns the mapping between convolutional filters and neighboring vertices dynamically using features generated by the network, which is closely related to the multi-head attention mechanism used in \cite{velivckovic2018graph}.

\mypar{Geometry-aware methods}
A number of methods have been developed that take the geometrical arrangement of vertices and faces explicitly into account to define network layers.
SpiralNet~\cite{gong2019spiralnet++,lim2018simple}  enumerates the neighboring vertices following randomly generated spiral patterns around the central vertex. 
MeshCNN~\cite{hanocka19tog} defines a convolution operation on edges aggregating information from their incident triangular edges and proposes a task-driven pooling operation based on the edge-collapse operation \cite{hoppe1997view}.  
An attention-based approach was explored in~\cite{milano2020primal}, which combines primal and dual graph representations. 
Their primal graph connects faces that share an edge, where the dual graph connects edges that are part of the same face. They use a pooling operation based on edge contraction on the mesh.
In contrast, we assume in our work that the vertices and edges of an input triangular mesh form the primal graph, and construct a dual mesh built on the faces. Rather than using a generic graph-based convolution on the dual mesh, we can therefore exploit the three-neighbor regularity to propose a dual mesh-based convolution. Additionally, we present an evaluation of different features defined on faces and examine the ability to learn connectivity-independent representations using different approaches.

%% file: sec_method.tex
\begin{figure*}
\centering
\includegraphics[width=0.7\textwidth]{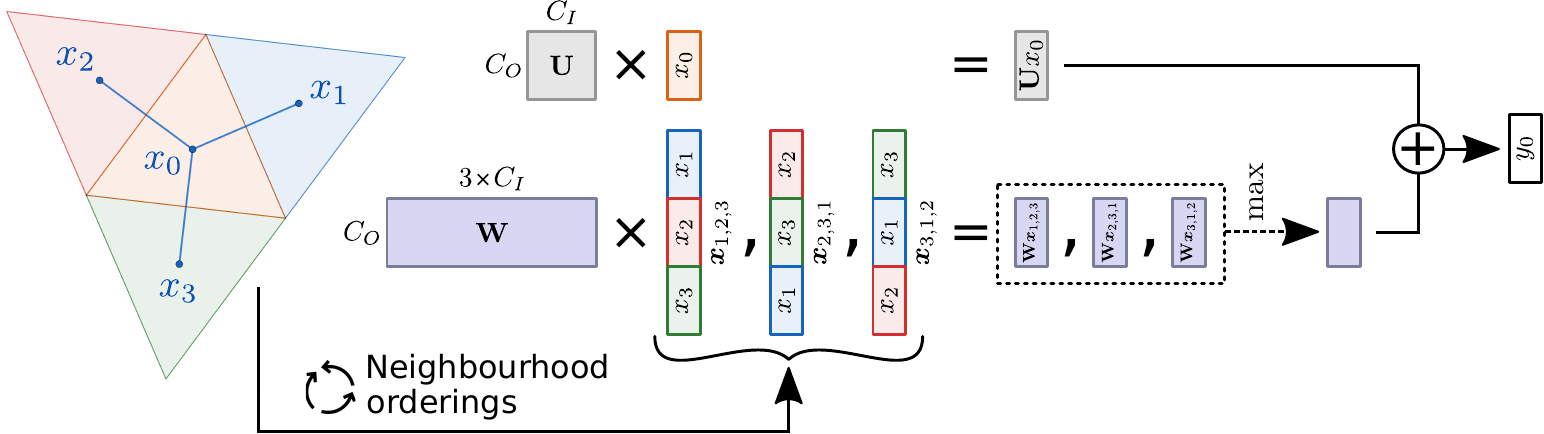} 
\hfill 
\includegraphics[width=0.55\columnwidth]{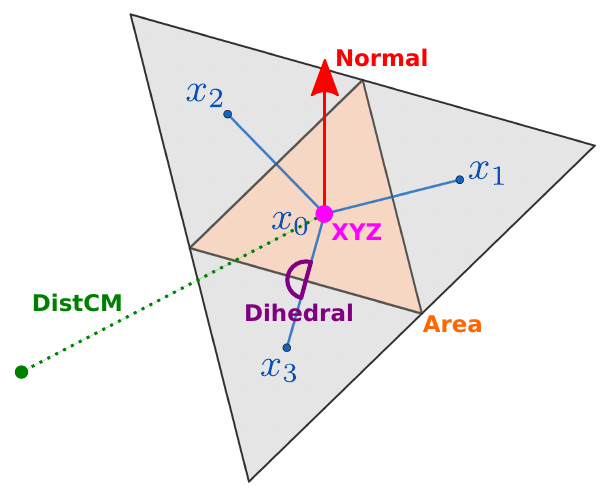}
\caption{
Left: Illustration of the DualConvMax layer that max-pools over different orderings of the local neighborhood. 
Right: Illustration of the triangular primal mesh $\mathcal{M}$ (in black) and the corresponding dual mesh $\mathcal{D}$ (in blue) with vertices $\{x_0, x_1, x_2, x_3\}$. Note that the central vertex $x_0$ of $\mathcal{D}$ has exactly three neighbors. We also illustrate the input features on the dual mesh that we consider in this work.
}
\label{fig:dual}
\label{fig:dual_conv_order}
\end{figure*}

\section{Method}
\label{sec:method}

Convolutional networks carry an inductive bias that meaningful information can be extracted from a consistent local neighborhood of the input, which is implemented using a set of fixed-sized trainable filters that are applied across the complete image. 
However, it is not obvious how to define such filters for meshes, due to their irregular local structure.
We address this difficulty by designing convolution on the dual of watertight triangular meshes, where each face has exactly three neighbors.

A primal mesh $\mathcal{M}$ is defined by $N_V$ vertices and $N_F$ faces. The dual $\mathcal{D}$ of $\mathcal{M}$ is defined as a mesh 
where each vertex is centered on a face of $\mathcal{M}$. 
These vertices in the dual $\mathcal{D}$ are connected by an edge if their corresponding two faces in the primal mesh $\mathcal{M}$ are adjacent. 
For a watertight triangular mesh $\mathcal{M}$,  each vertex in the dual $\mathcal{D}$ has exactly three neighbors by construction, while
in the primal vertices can have different numbers of neighbors.
In cases where the mesh $\mathcal{M}$ is not watertight, we can use zero-padding to ensure that every vertex in $\mathcal{D}$ has three neighbors \nitika{ or if the mesh is non-manifold, we remove the particular vertices. }
We note that in general, this approach can be extended to any $N$-edged polygonal mesh, where the face-based dual mesh will form a regular $N$-neighbor structure.

Below, we describe the two main building blocks of our networks: a dual convolutional layer tailored explicitly to the fixed 3-neighborhood and a dual to primal feature transfer layer. 
Finally, we describe the different input features defined over faces that we consider in our experiments.

\subsection{Dual convolution}
\label{ssec:dual_conv}

Given a face in $\mathcal{M}$, represented by $x_0$ in Figure \ref{fig:dual_conv_order}, we wish to define the convolution as the dot product of the weights with the features of the neighbors, similar to a convolutional layer over regular pixel grids. 
Although the neighbors of a face can be assigned a unique clockwise orientation defined \wrt the central face normal, their order (\ie which neighbor comes first) is not unique. 
To resolve the ordering ambiguity for the neighboring faces, we use a strategy analogous to angular max-pooling~\cite{masci15iccvw}.
%
%
Let $C_I$ and $C_O$ denote the number of input and output feature channels, respectively. 
The central node's feature $x_0$ is always multiplied with the same weights $\mathbf{U}\in\R^{C_O\times C_I}$. Weights $\mathbf{W}\in\R^{C_O\times 3C_I}$ are applied to the local neighbors using their three possible orderings, followed by a coordinate-wise max-pooling across the orders:
\begin{equation}
y_0 = \mathbf{U} x_0  + \max\{ \mathbf{W} \boldsymbol{x}_{1,2,3}, \mathbf{W} \boldsymbol{x}_{2,3,1}, \mathbf{W} \boldsymbol{x}_{3,1,2}\},
\label{eq:dual_conv_max}
\end{equation}
where $y_{0}\in\R^{C_O}$ is the output feature, $\boldsymbol{x}_{1,2,3}\in\R^{3C_I}$ denotes the concatenation of the neighbors' features $x_1$, $x_2$ and $x_3$ in this order. 
We refer to this layer as \textbf{DualConvMax}. 
See Figure \ref{fig:dual_conv_order} for an illustration.

\subsection{Dual to primal feature transfer}
\label{ssec:dual_to_primal}

To handle  cases where the prediction targets and/or the ground-truth for training are defined only on the vertices of the primal mesh, 
we define a {\bf Dual2Primal} layer to transfer the features from the dual back to the original mesh. The features transferred to the primal mesh can then be used to measure the training loss or make predictions for evaluation.

Given a mesh $\mathcal{M}$, 
we construct a vertex-face adjacency matrix $\mathbf{A}\in \R^{N_V\times N_F}$, and derive the vertex-degree matrix $\mathbf{D}=\textrm{diag}( \mathbf{A}\boldsymbol{1}_{N_F} )$, where $\boldsymbol{1}_{N_F}$ is a vector of ones of size $N_F$. 
The diagonal of $\mathbf{D}$ contains for each vertex in the primal mesh the number of faces to which it belongs.
The output features $\mathbf{F}_{Dual}$ of the dual neural network are converted into features $\mathbf{F}_{Primal}$ on the primal mesh by averaging for each vertex the features of all faces incident to that vertex:
\begin{equation}
    \mathbf{F}_{Primal} = \mathbf{D}^{-1}\mathbf{A} \mathbf{F}_{Dual}.
    \label{eq:dualtooriginal}
\end{equation}
We then apply the loss defined on the primal mesh and back-propagate the gradients through the dual network.

\jkb{It is interesting to consider alternative dual-to-primal conversion schemes, \eg based on the local geometry or attention mechanisms, 
but we leave this for future work.
}

\subsection{Input  features from dual mesh}
\label{ssec:features}

Using faces rather than vertices as inputs for our deep network allows the use of features that are naturally defined over faces but not over vertices.  
In our experiments, we explore the effectiveness of the following 
input features defined over faces: 
(i)  \textbf{XYZ}: the coordinates of the center of mass of the face. 
(ii)  \textbf{Normal}: the unit vector in the direction of the face normal.
(iii)  \textbf{Dihedral}: the angles (in radians) between the face and its neighbors. 
(iv) 
\textbf{Area}: the surface area of the face.
(v) \textbf{DistCM}: the Euclidean distance between the center of mass of the full mesh and the face. 
We illustrate these features in \fig{dual}. They offer  
different degrees of invariance; ranging from XYZ that does not offer any invariance, to dihedral angles which are invariant to translation, rotation, and scaling of the 3D shape. 
We note that the dihedral angles are defined per adjacent face, so we use them  by setting $x_0 = 0$ and $x_i=\text{Dihedral}_{0,i}$ in Equation \ref{eq:dual_conv_max}.
The remaining features are defined per face, we can directly use them as inputs proper to each face. We also consider combinations of these features by concatenating them into a larger input feature. 





%% file: sec_experiments.tex
\section{Experimental evaluation}
\label{sec:results}

We first describe our experimental setup in Section \ref{ssec:experimental_setup}. We then present our experimental results when training our models on the Faust-Synthetic and Faust-Remeshed datasets in sections \ref{ssec:experiments_faust} and \ref{ssec:remeshed} respectively.

\subsection{Experimental setup}
\label{ssec:experimental_setup}

We closely follow the experimental setup of previous work~\cite{fey18cvpr,monti17cvpr,verma18cvpr}, and 
perform evaluations on the Faust human shape dataset~\cite{bogo14cvpr}.
It consists of 100 watertight triangular meshes with ten subjects, each striking ten different poses; the first 80 meshes are used for training and the last 20 meshes for testing. 
The meshes in this dataset are obtained by fitting a fixed template mesh with 6,890 vertices and 13,776 faces to raw scan data.   
We refer to this dataset as {\bf Faust-Synthetic} in the evaluations.
All meshes have the same underlying connectivity, and the ground-truth is defined by a one-to-one correspondence of the vertices.

\begin{figure}
    \centering
	\begin{tabular}{cc}
	\includegraphics[scale=0.02]{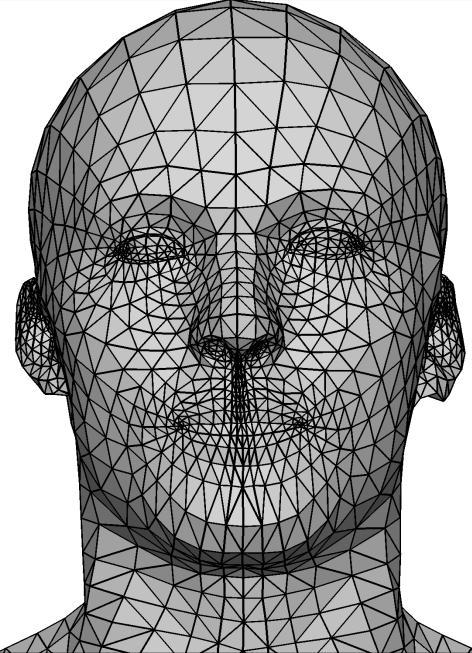} &
	\includegraphics[scale=0.02]{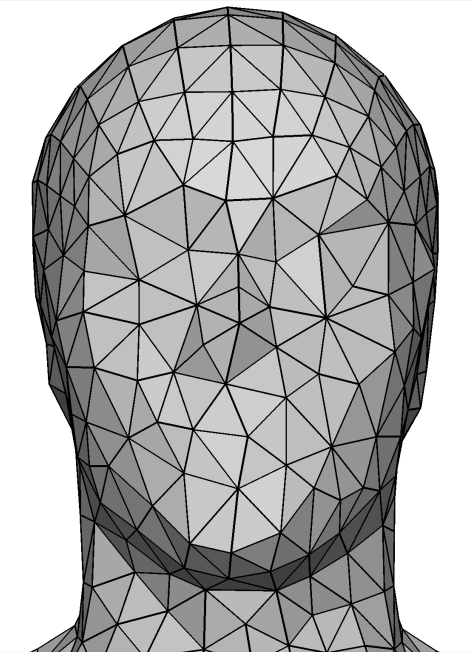}
	\\
	\textbf{Faust-Synthetic} & \textbf{Faust-Decimated (50\%)}
	\\
	\includegraphics[scale=0.02]{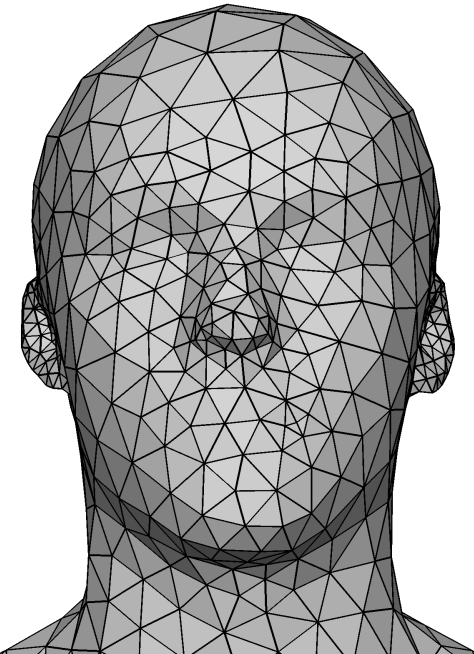} &
	\includegraphics[scale=0.02]{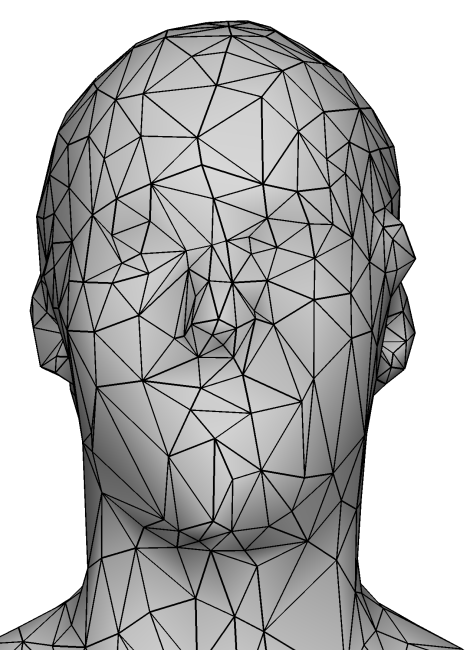}
	\\
	\textbf{Faust-Remeshed} & \textbf{Faust-Scan}
	\\
	\end{tabular}
\caption{Visualization of a mesh from the template-fitted Faust dataset, decimated by 50\%, re-meshed version and original scan.
}
\label{fig:faust_data}
\end{figure}

\begin{figure*}
\centering
\resizebox{\textwidth}{!}{
	\includegraphics{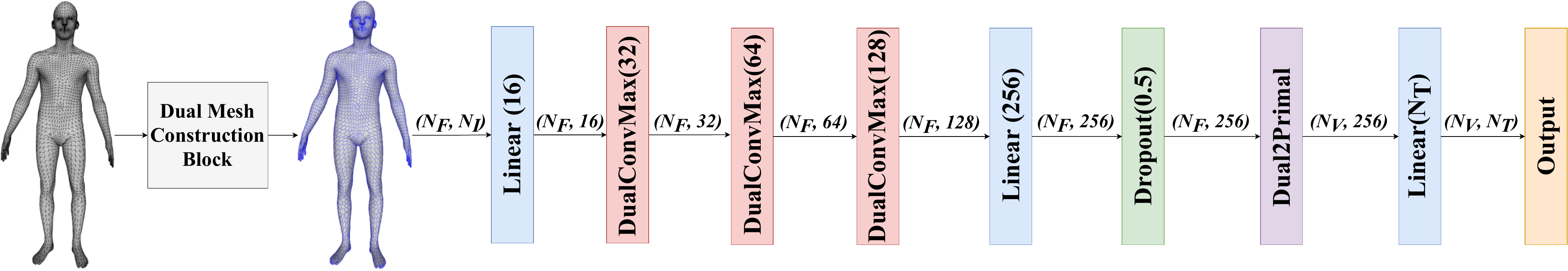}
}
\caption{Architecture for our dual mesh networks. For FeaStNet--Dual  we replace  DualConvMax with FeaStConv.
}
\label{fig:dual_arc}
\end{figure*}

To allow for more challenging evaluations with varying mesh topologies, we consider three other versions of the Faust dataset, see  \fig{faust_data}:
\begin{itemize}
  \setlength\itemsep{0em}
    \item \textbf{Faust-Decimated}: We use quadric edge collapse~\cite{garland1997surface} to reduce the resolution of the meshes in Faust-Synthetic by up to 50\%. 
    While mesh decimation is a fairly straightforward way to assess robustness to changes in the mesh structure, we note that it changes some parts of the mesh more drastically than others.
    
    \item \textbf{Faust-Remeshed}: We consider the re-meshed version of the dataset from~\cite{ren2018continuous} as a more realistic and challenging testbed. 
    It was obtained by re-meshing every shape in the Faust-Synthetic dataset independently using the LRVD method~\cite{yan2014low}. 
    Each mesh in the resulting dataset consists of around 5,000 vertices and has a unique mesh topology. 
    While offering an interesting testbed, the re-meshed data does not come with dense one-to-one vertex ground-truth correspondence. A partial ground-truth is however available for roughly 3,500 vertices. 
    
    \item \textbf{Faust-Scan}: We consider the raw scan data that underlies the dataset. 
    It contains 200 high-resolution meshes, with the same 10 people striking 20 different poses.
    The average number of vertices in each scan is around 172,000, which we reduce 
    using quadric edge collapse decimation~\cite{garland1997surface} to bring closer to the reference template with 6,890 vertices.
    We note that this dataset is very challenging as it does not contain watertight meshes and all meshes have different topologies. 
    There is no ground-truth available, so we only perform a qualitative evaluation on this version of the dataset.

\end{itemize}


\mypar{Network architectures and training}
\fig{dual_arc} describes the dual mesh-based architecture that we use in our experiments, where $N_V$ and $N_F$ are the number of vertices and faces in the original primal mesh respectively,  $N_T$ the number of target labels and $N_I$ the number of input features. We use ``Linear($K$)'' to indicate fully connected layers, 
and ``DualConvMax($K$)'' to indicate graph-convolutional layers (defined in \ssect{dual_conv}), producing each $K$ output feature channels. ``($N$, $K$)'' denotes feature maps of size $N$ and dimension $K$. 
We apply the Exponential Linear Unit (ELU) non-linearity \cite{ClevertUH15} after every DualConvMax layer and every linear layer, except for the last one. 
We also indicate the  rate for the Dropout layer~\cite{srivastava14jmlr}. 

Similar to previous work~\cite{fey18cvpr,monti17cvpr,verma18cvpr}, we formulate the shape correspondence task as a vertex labeling problem, where the labels are the set of vertices in a given reference shape. We implement our method using the PyTorch Geometric framework~\cite{Fey2019pyg}, and train models using the Adam optimizer~\cite{kingma15iclr} to minimize the cross-entropy classification loss.
Additional details on the training can be found in the supplementary material.

\jkb{The receptive field of the primal and dual architectures grows at the same rate when adding  layers, because in both cases new elements (vertices or faces) are within one edge distance.
Our networks rely on fairly local information, using three DualConvMax layers, and one Dual2Primal layer.
}

\mypar{Evaluation metrics}
Following previous work~\cite{fey18cvpr,monti17cvpr,verma18cvpr}, we report the accuracy, \ie the fraction of vertices for which the exact correspondence has been correctly predicted.
In addition, we report the mean geodesic error, \ie the average of the geodesic distance between the ground-truth and the prediction, normalized by the geodesic diameter and multiplied by 100. 
We believe the mean geodesic error metric is more informative than the accuracy as a single-number comparison for the correspondence task. Rather than just counting the number of imperfect correspondences, it  considers how large these errors are.
\nitika{In particular, some methods may have a lower accuracy than others, but make fewer mistakes with large geodesic errors, leading to a smaller average geodesic error.}




\subsection{Results with training on Faust-Synthetic }
\label{ssec:experiments_faust}

The shape correspondence task on the Faust dataset is defined on the mesh vertices.
In our first experiment, we validate the use of the dual mesh to establish shape correspondence and the effectiveness of networks built on our DualConvMax and Dual2Primal  operators.
For this purpose, we use the XYZ position of the face centers as input and compare results to those obtained with FeaStNet~\cite{verma18cvpr} on the primal mesh. 
Since FeaStNet is a generic graph convolution method, it can be readily applied to the dual mesh. We refer to the results obtained using this approach as FeaStNet--Dual. This allows us to separate the effects of using the primal \vs dual mesh from the use of our DualConvMax layers.

\begin{table}[]
    \centering
    \resizebox{\columnwidth}{!}{
    \begin{tabular}{lllcc}
        \toprule
        Mesh & Method & Input & Geo. Err. & Accuracy  \\
        \midrule
        Primal & FeaStNet~\cite{verma18cvpr} & XYZ & $1.39$ & $88.1\%$ \\
        \midrule
        \multirow{2}{*}{Dual} & FeaStNet--Dual & XYZ & $0.18$ & $92.7\%$ \\
         & DualConvMax & XYZ & $0.26$ & $95.5\%$ \\
        \bottomrule
    \end{tabular}
    }
    \caption{Mean geodesic error and correspondence accuracy. Comparing primal and dual methods on the Faust-Synthetic dataset. 
    }
    \label{tab:faust_dual}
\end{table}


\begin{figure}
	\centering
	\resizebox{0.8\columnwidth}{!}{
	\begin{tabular}{c}
	    \includegraphics[scale=1]{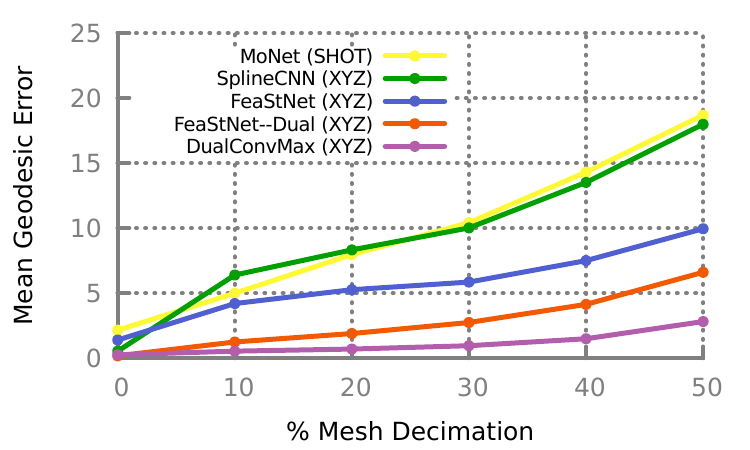} \\
	    \includegraphics[scale=1]{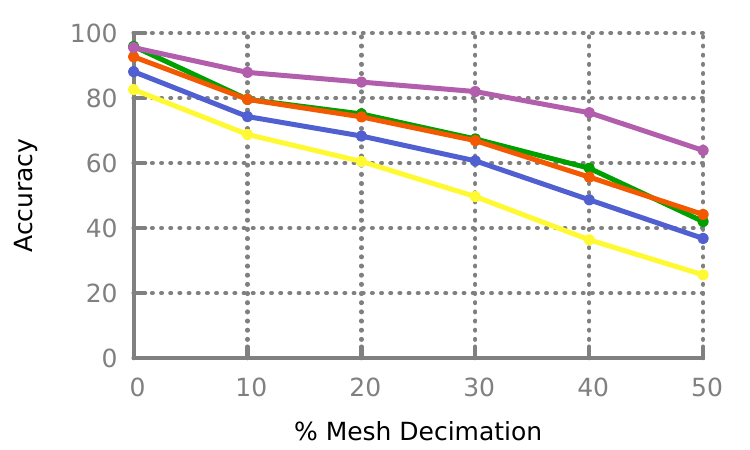}
	\end{tabular}
	}
\caption{Mean geodesic error and accuracy for Faust-Decimated test meshes with XYZ. All methods are trained on the original full resolution Faust-Synthetic meshes. 
}
\label{fig:faust_decimated_sota_errors}
\end{figure}

\begin{figure*}
	\setlength{\tabcolsep}{15pt}
    \centering
	\begin{tabular}{cccccc}
	 &
	\includegraphics[height=90pt]{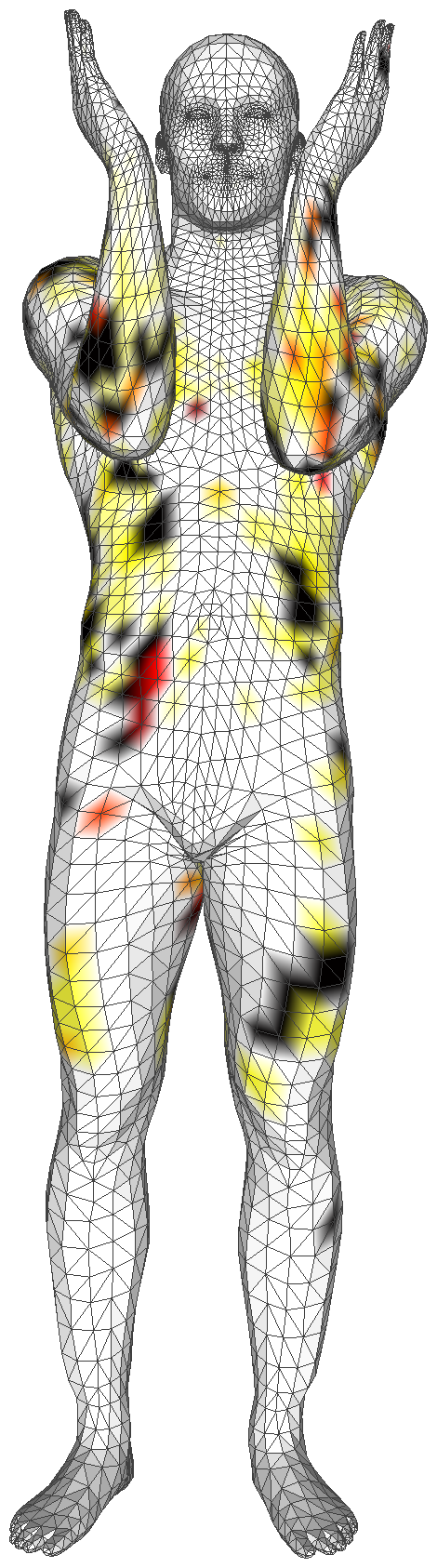} &
	\includegraphics[height=90pt]{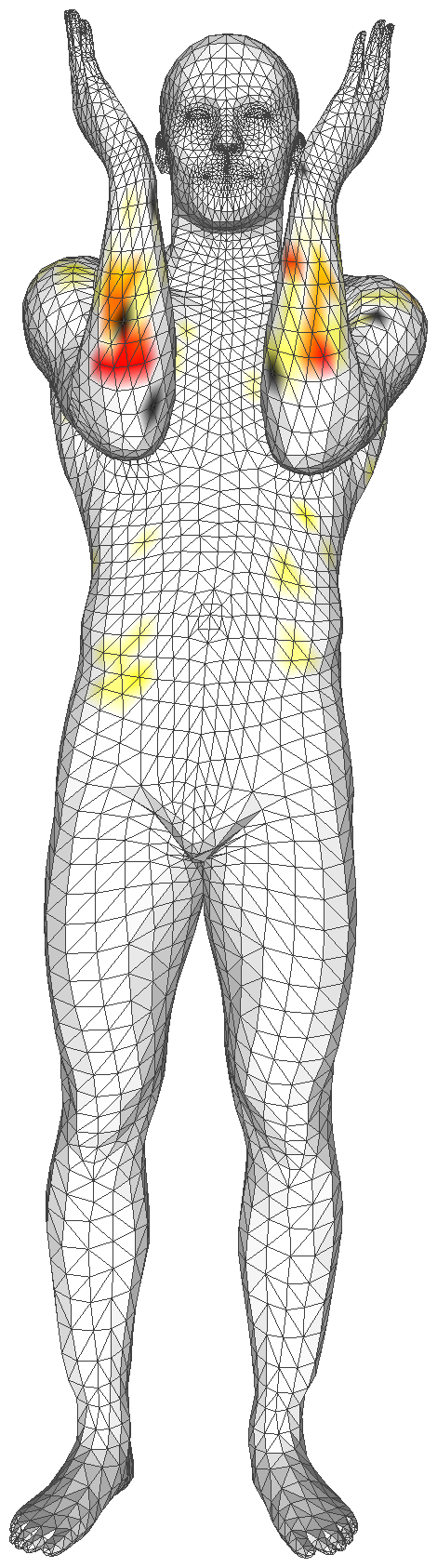} &
	\includegraphics[height=90pt]{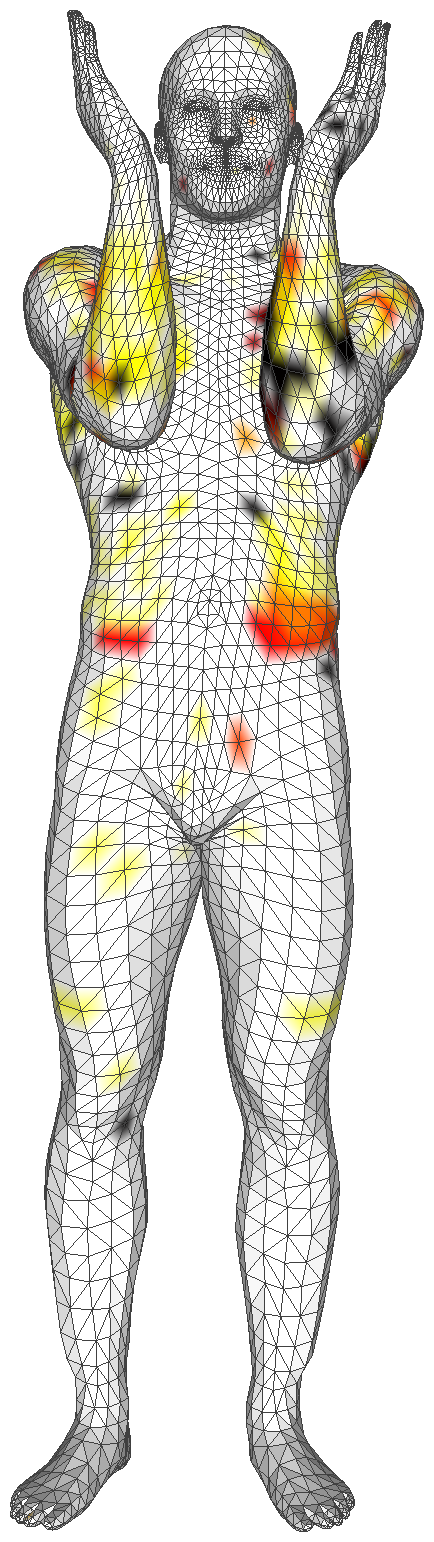} &
	\includegraphics[height=90pt]{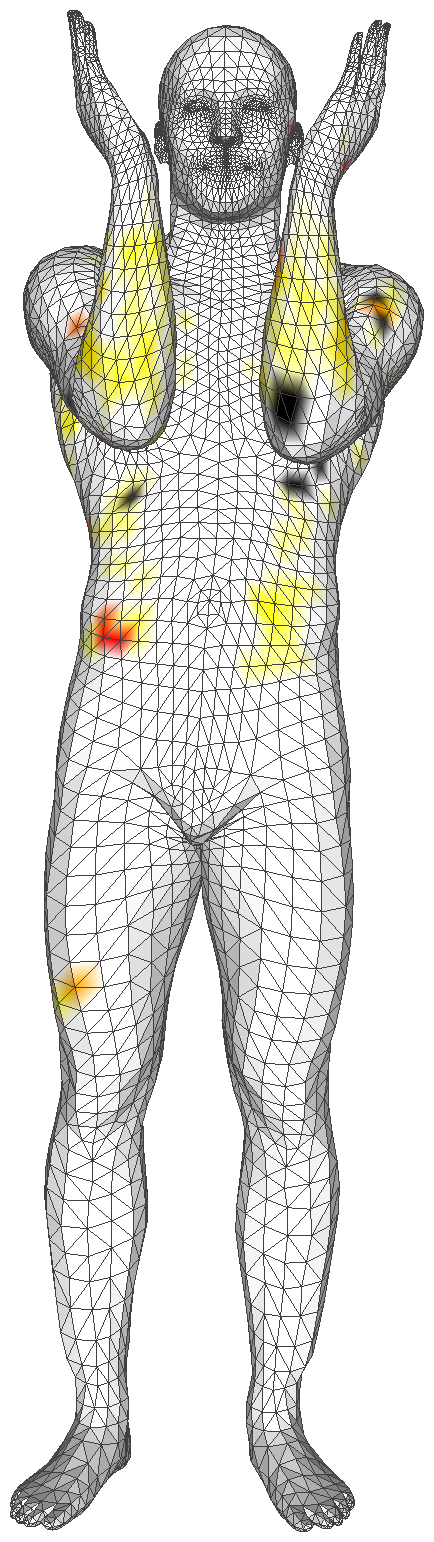} &
	\includegraphics[height=90pt]{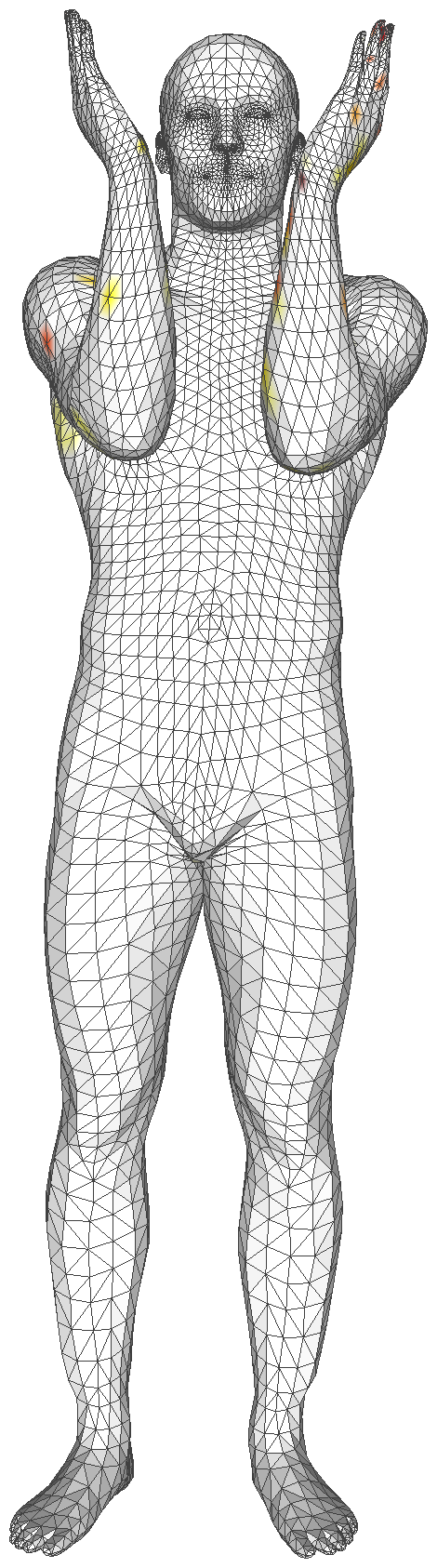}
	\\ 
	\includegraphics[height=60pt]{figures/errors_colormap} &
	 \includegraphics[height=90pt]{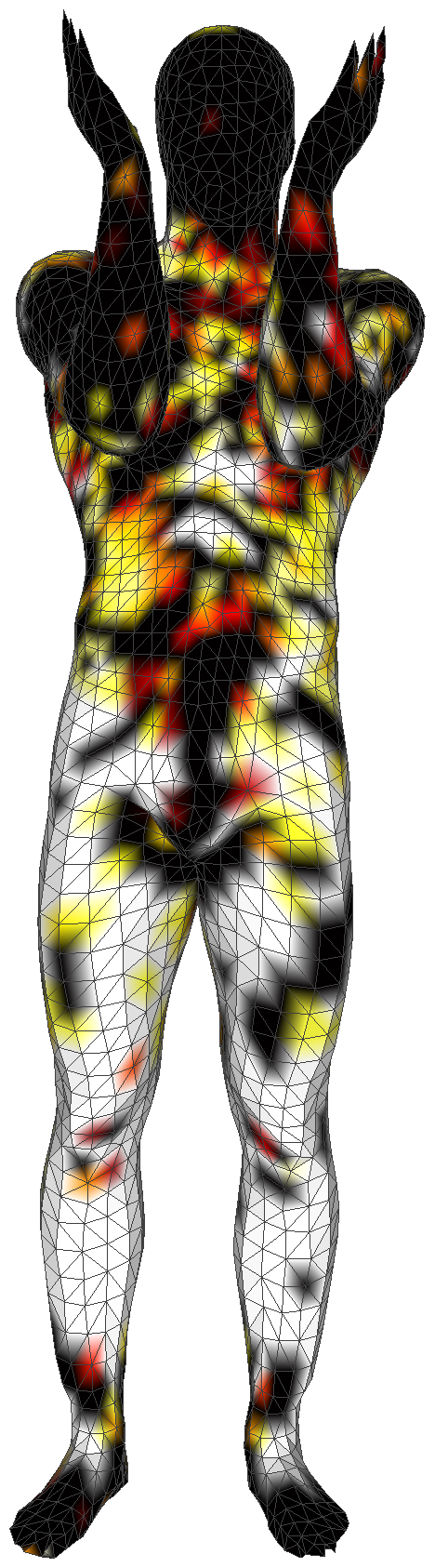} &
	 \includegraphics[height=90pt]{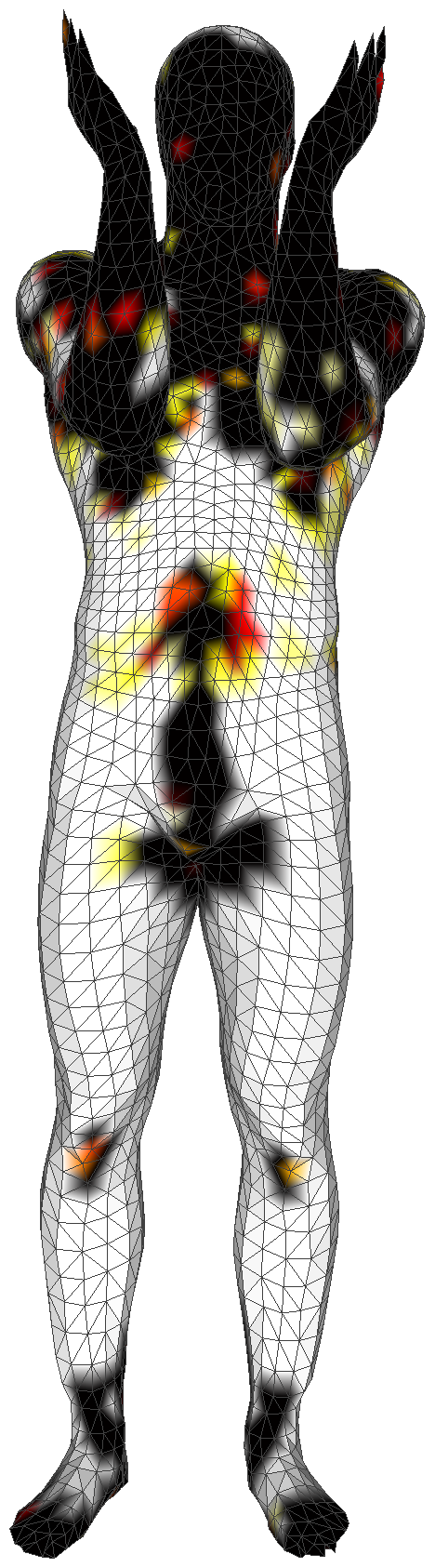} &
	\includegraphics[height=90pt]{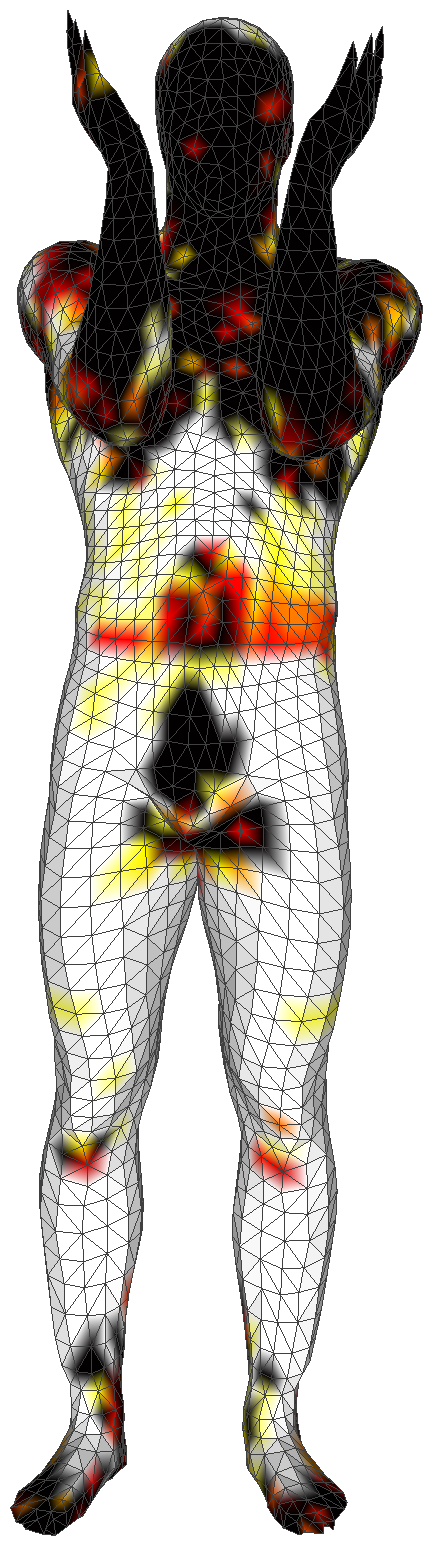} &
	\includegraphics[height=90pt]{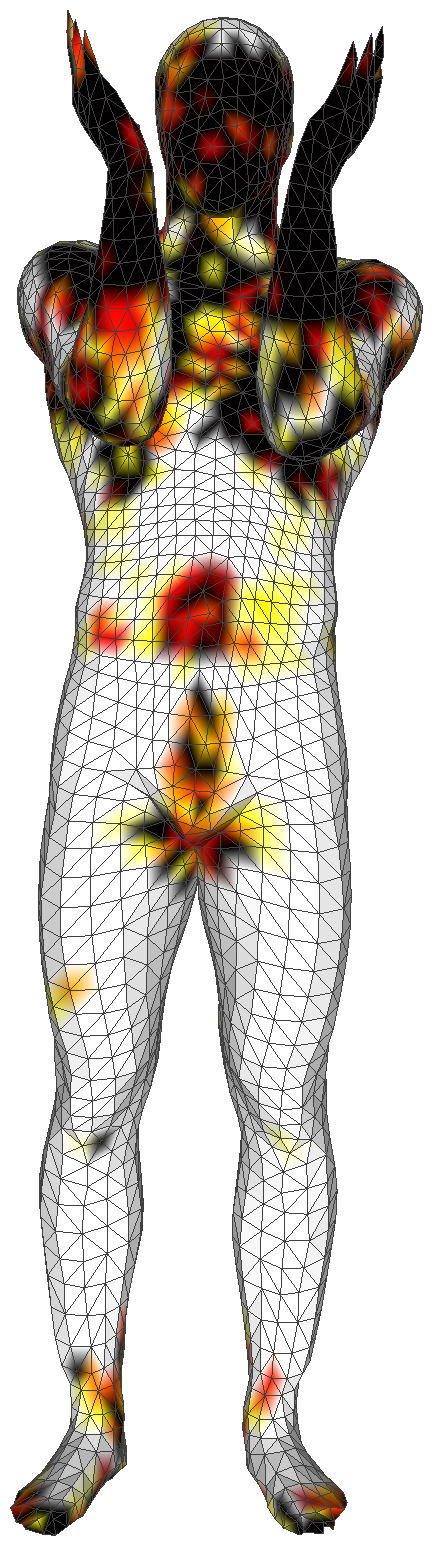} &
	\includegraphics[height=90pt]{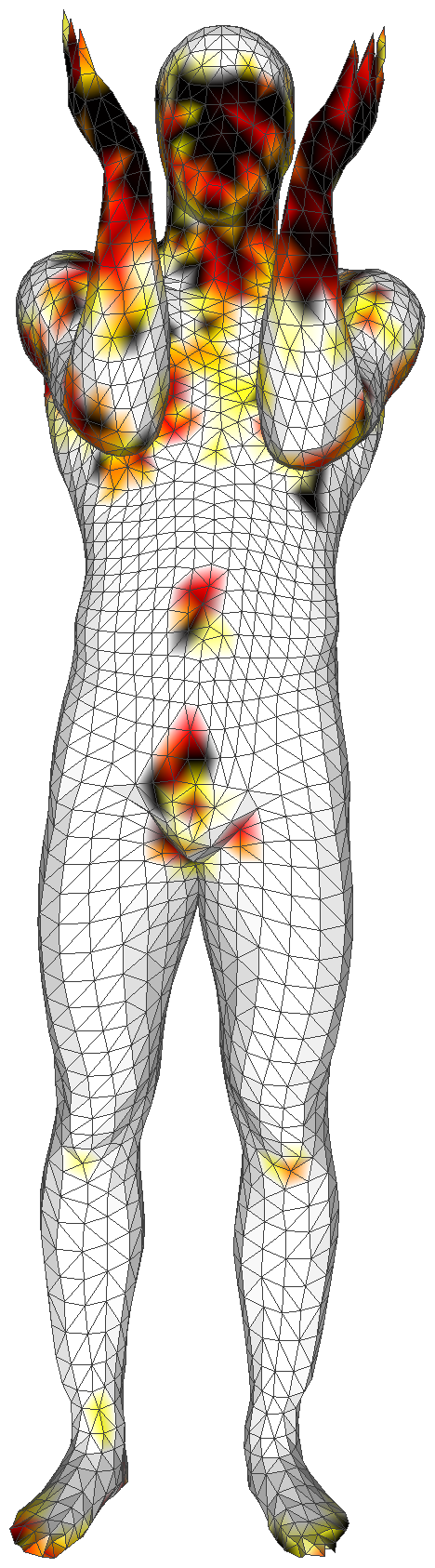}
	\\
	 & \footnotesize{\textbf{MoNet}} & \footnotesize{\textbf{SplineCNN}} & \footnotesize{\textbf{FeaStNet}} & \footnotesize{\textbf{FeaStNet--Dual}} & \footnotesize{\textbf{DualConvMax}}
	\\
	 & \footnotesize{(SHOT)} & \footnotesize{(XYZ)} & \footnotesize{(XYZ)} & \footnotesize{(XYZ)} & \footnotesize{(XYZ)}
	\\
	\end{tabular}
\caption{Visualizations of geodesic correspondence errors for a full resolution Faust-Synthetic test mesh (top row), and the same mesh decimated by 50\% (bottom row) of the Faust-Decimated dataset. Models are trained on the full resolution Faust-Synthetic meshes.
}
\label{fig:faust_decimated_errors}
\end{figure*}

\mypar{Correspondences on Faust-Synthetic}
We present the results in \tab{faust_dual}.
Comparing FeaStNet and FeaStNet--Dual, we observe that the Dual2Primal layer successfully transfers features learned over the faces to the primal vertices. Moreover, using the dual mesh improves performance: the mean normalized geodesic error drops from $1.39$ to $0.18$, and the accuracy increases from $88.1\%$ to $92.7\%$.
Next, we observe that our DualConvMax performs better than FeaStNet while obtaining the highest overall accuracy (95.5\%). 
Note that both dual-based approaches are better than FeaStNet in terms of accuracy and obtain much lower mean geodesic errors. 

Based on these encouraging results, we now turn to evaluations in more challenging conditions. 
The Faust-Synthetic dataset is unrealistic in that all meshes share one identical mesh structure connectivity.
Therefore, it is possible that deep networks that are trained on them learn to exploit this property to solve the correspondence problem on this dataset without being able to generalize to shapes with other mesh topologies.
To assess to what extent this happens, 
in the experiments below, we train the networks on Faust-Synthetic and test the resilience to connectivity changes on Faust-Decimated.

\mypar{Transfer from Faust-Synthetic to decimated meshes}
We compare our approach with the previous state-of-the-art methods in \fig{faust_decimated_sota_errors}.
We observe that the networks that use the dual mesh are more robust to connectivity changes than MoNet, SplineCNN, and FeaStNet, which are based on the primal mesh.
Our DualConvMax improves the accuracy by $2.8\%$ as compared to FeaStNet applied to the dual (FeaStNet--Dual) in the case without connectivity changes, and leads to substantially better accuracy of $63.9\%$ compared to the $44.2\%$  when meshes are decimated by $50\%$.
We note that the methods on the primal mesh all achieve poor mean geodesic errors on the decimated meshes.
Considering the results obtained with FeaStNet--Dual, we note that the improved performance of DualConvMax \wrt previous methods (MoNet, SplineCNN, and FeaStNet) is both due to the use of the dual mesh structure and to the DualConvMax operator that we specifically designed for the dual mesh.

\begin{figure*}
	\resizebox{\textwidth}{!}{
	\begin{tabular}{cccccc}
	\includegraphics[height=90pt]{orig_scans/faust_ref} &
	\includegraphics[height=90pt]{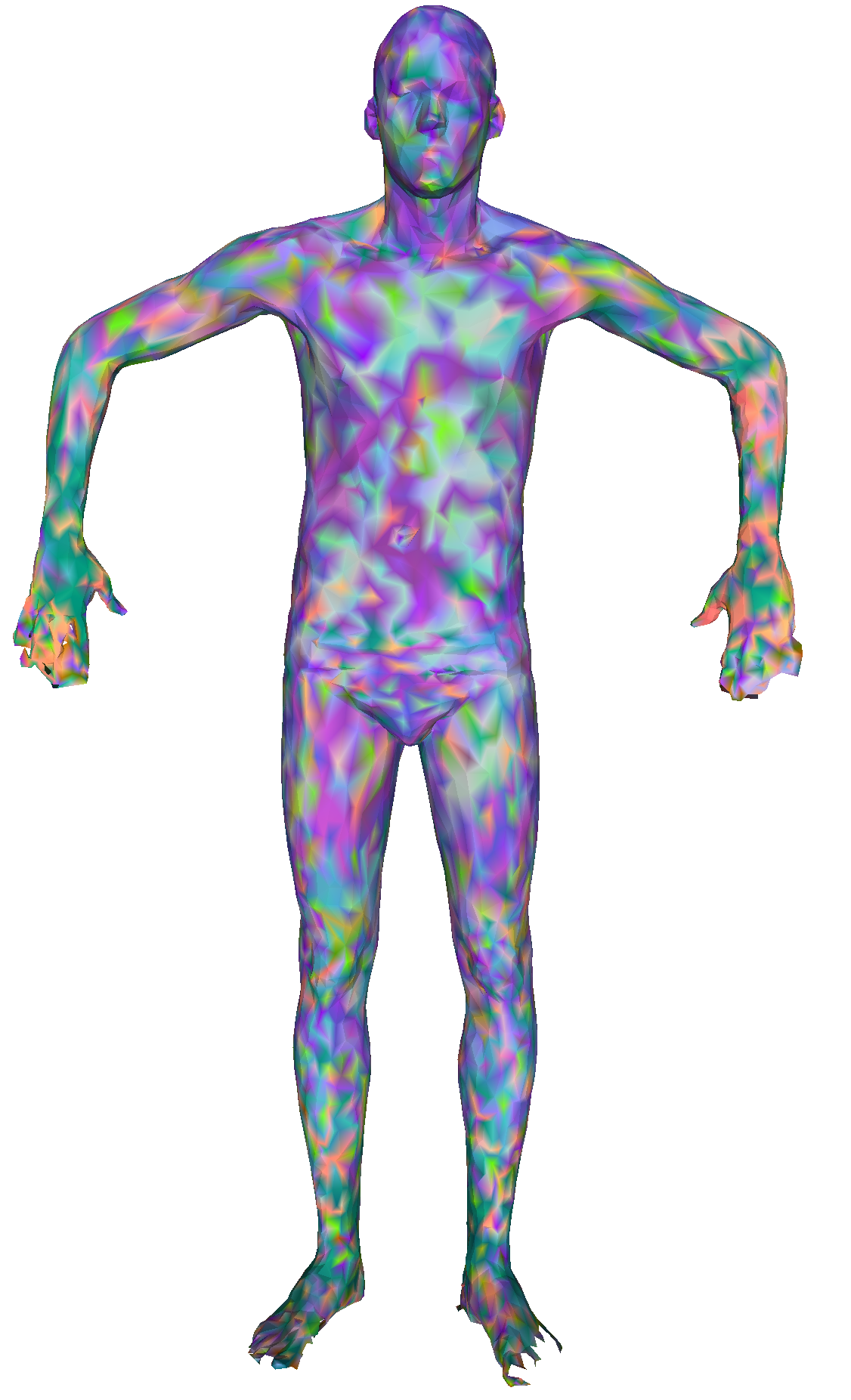} &
	\includegraphics[height=90pt]{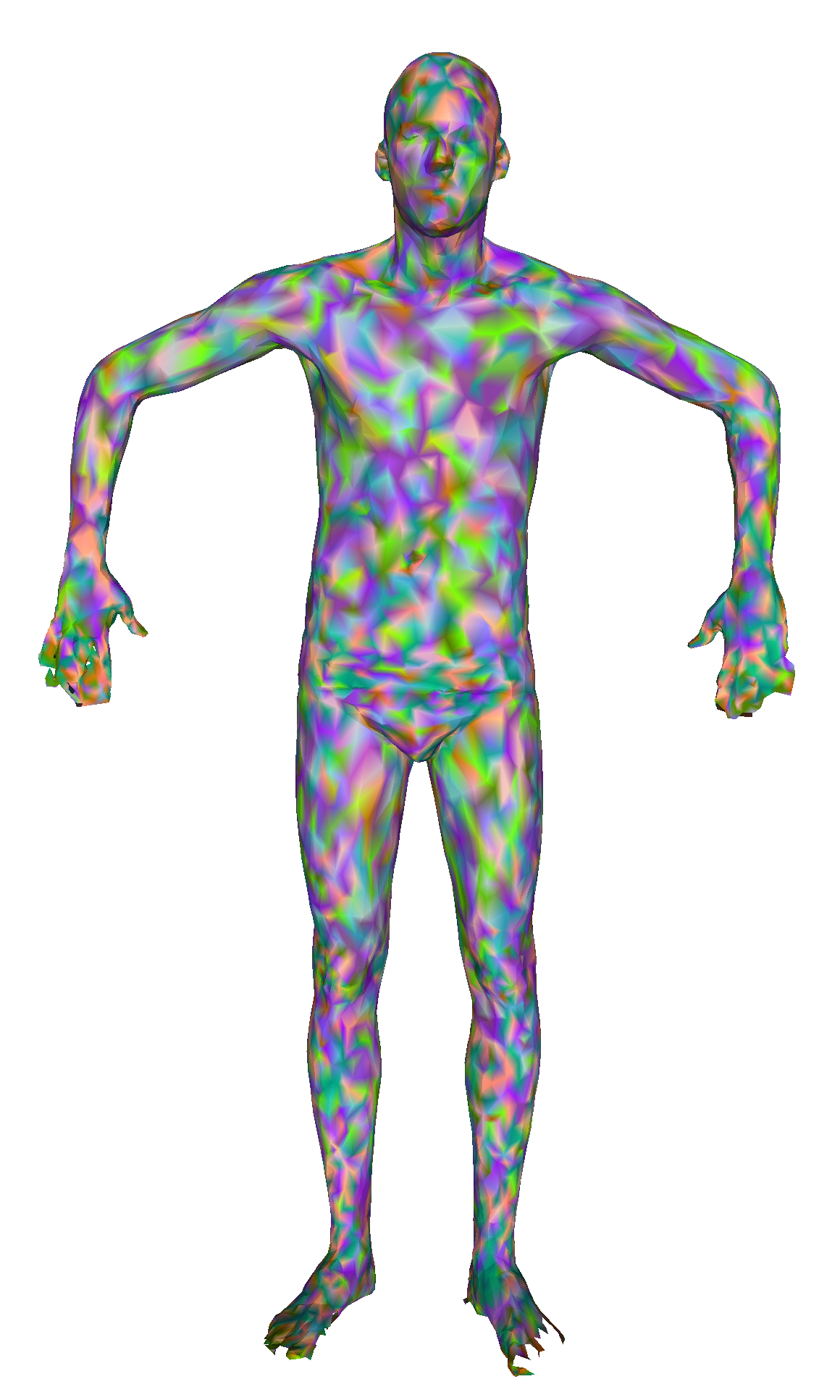} &
	\includegraphics[height=90pt]{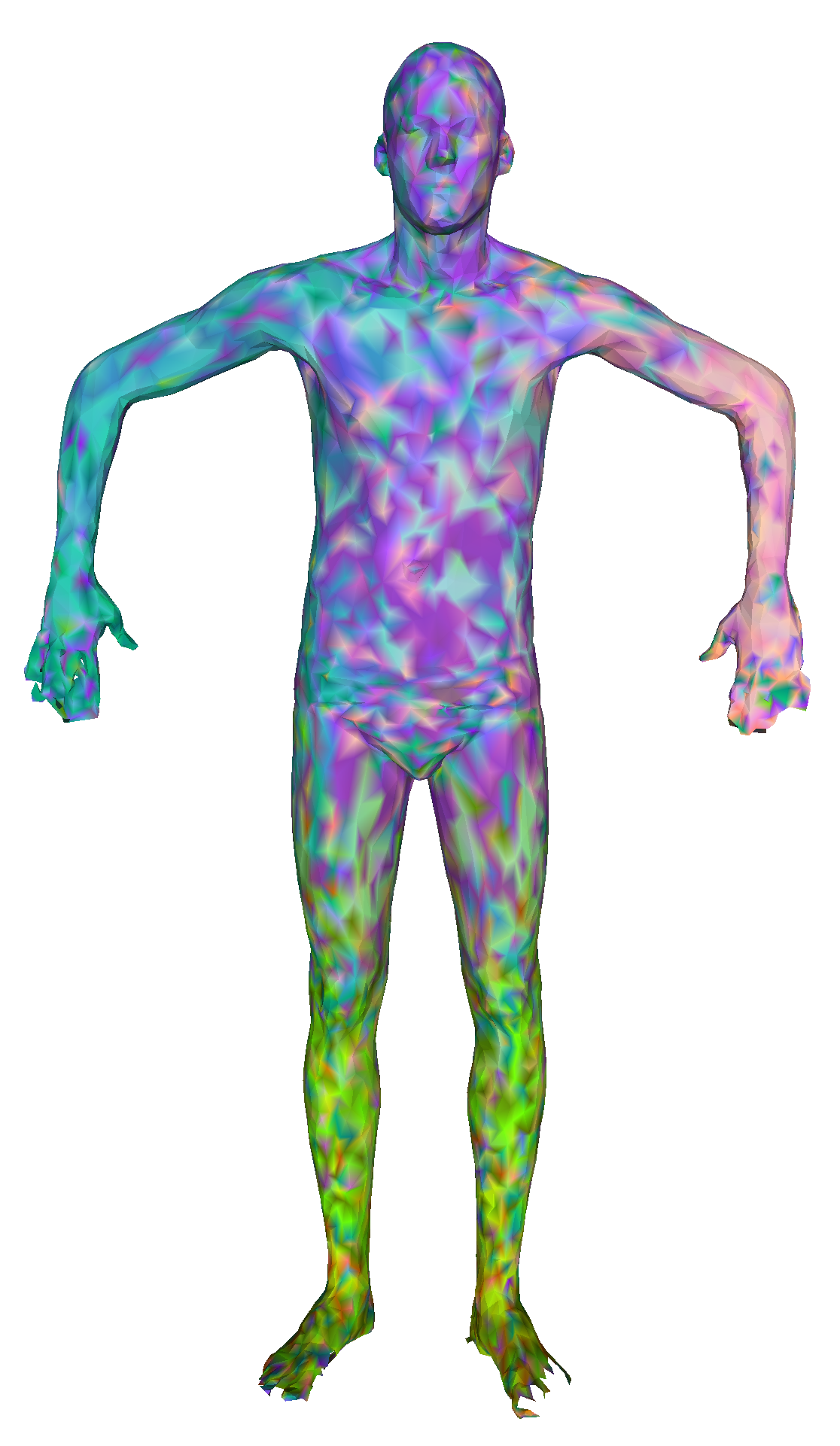} &
	\includegraphics[height=90pt]{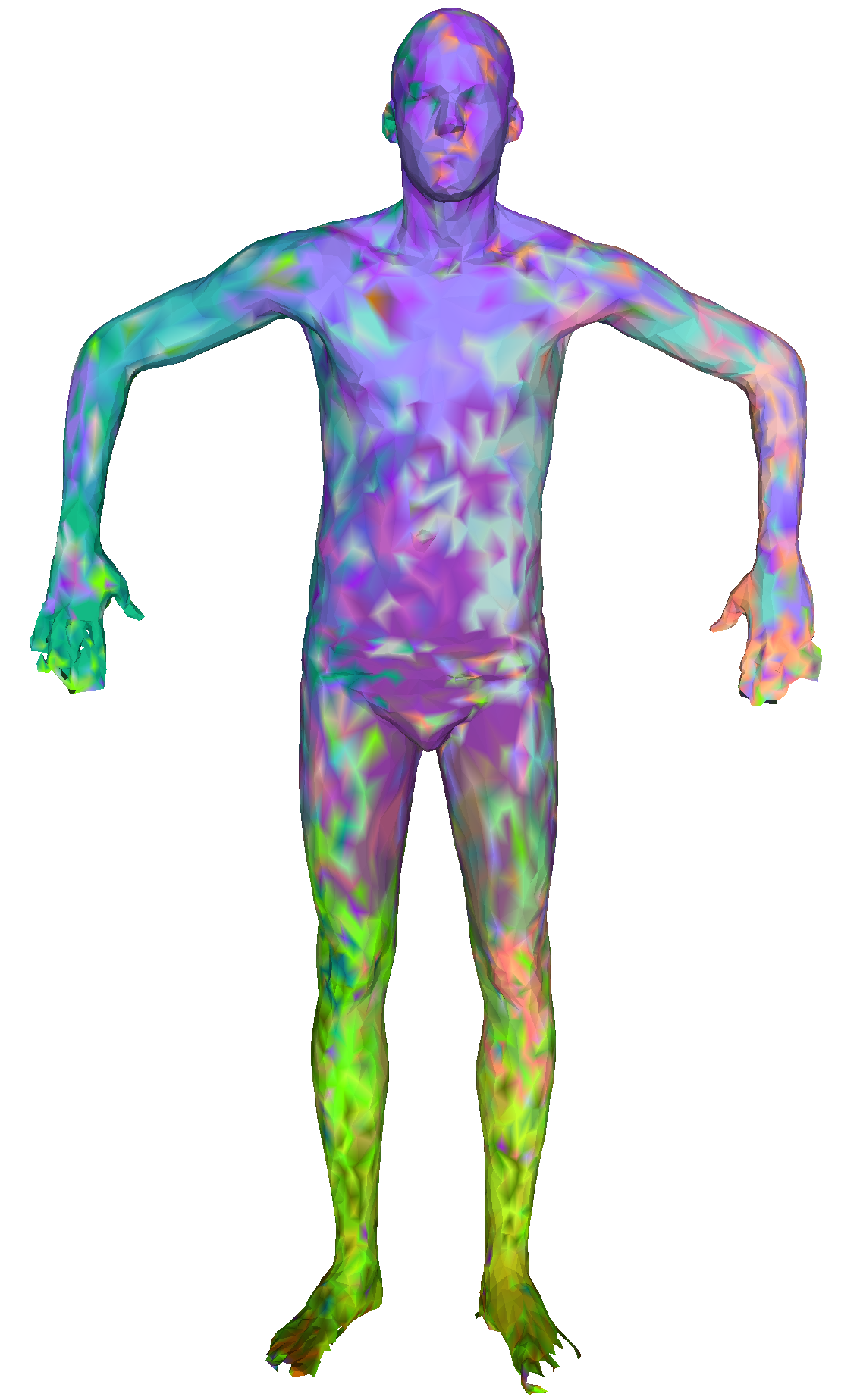} &
	\includegraphics[height=90pt]{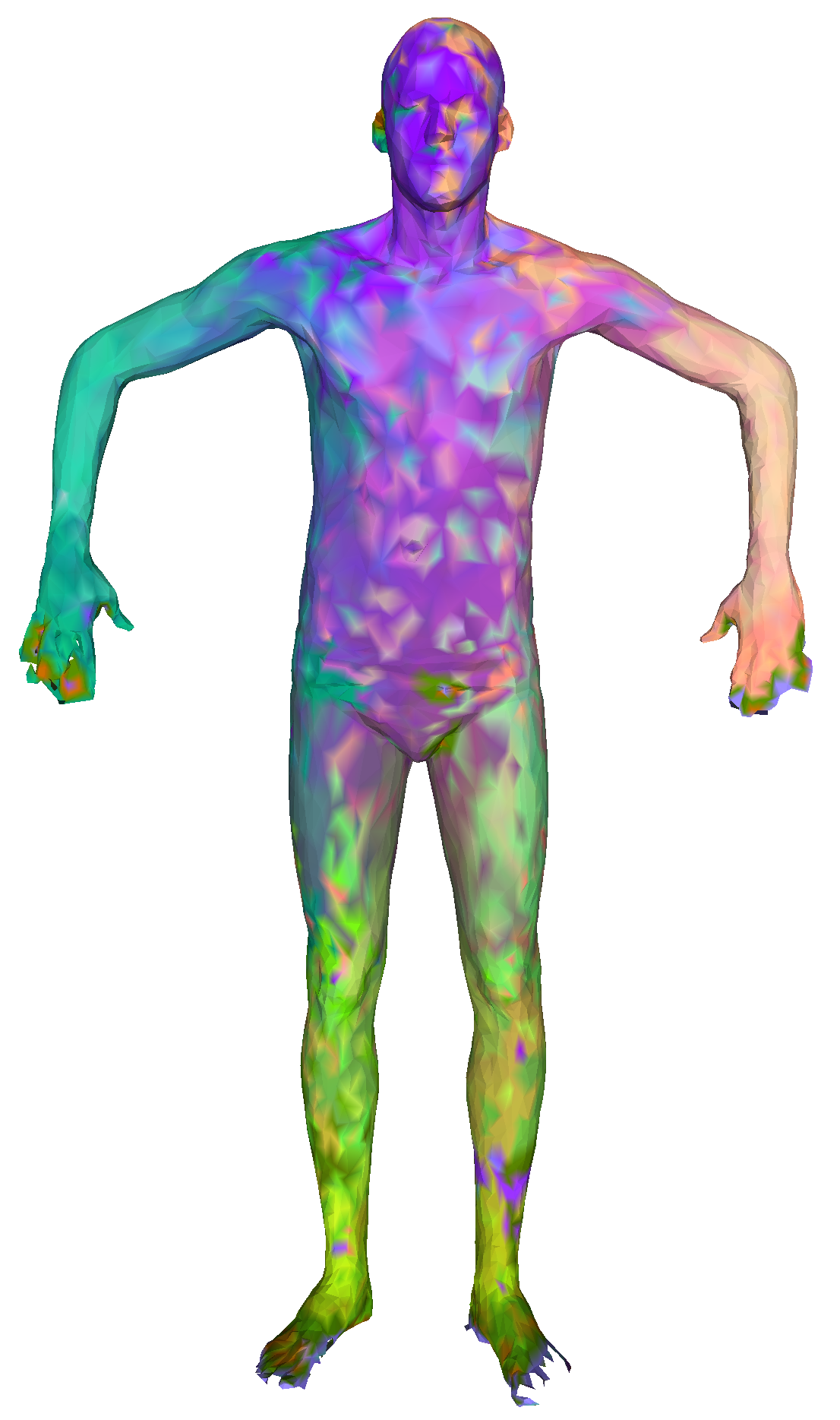}  
	\\
	\footnotesize\textbf{Faust-Synthetic} & \footnotesize\textbf{MoNet} & \footnotesize\textbf{SplineCNN} & \footnotesize\textbf{FeaStNet} & \footnotesize\textbf{FeaStNet--Dual} & \footnotesize\textbf{DualConvMax}
	\\
	\footnotesize\textbf{Reference} & \footnotesize{(SHOT)} & \footnotesize{(XYZ)} & \footnotesize{(XYZ)} & \footnotesize{(XYZ)} & \footnotesize{(XYZ)}
	\\
	\end{tabular}
}
\caption{Visualizations of texture transfer for Faust-Scan test results.
All methods have been trained on the Faust-Synthetic dataset.}
\label{fig:test_scan_train_faust}
\label{fig:test_remeshed_train_faust}
\end{figure*}

We qualitatively compare the results of MoNet, SplineCNN, and FeaStNet on the primal, FeaStNet--Dual and our DualConvMax in terms of geodesic errors in \fig{faust_decimated_errors}. We provide  an example of a non-decimated test mesh and its 50\% decimated version.
We observe marked improvements in the results on the decimated mesh by using the dual rather than primal mesh and further substantial improvements by using our DualConvMax approach rather than FeaStNet--Dual.
This confirms what was observed in terms of accuracy and mean geodesic error before.
We provide more qualitative results for this experiment in the supplementary material.

\mypar{Qualitative results on Faust-Scan}
Above we observed that the approaches based on the dual mesh are more robust to topological changes induced by mesh decimation. 
We now turn to a qualitative evaluation on the Faust-Scan 
dataset. 
In this dataset, 
the topological changes appear across the entire shape, where the mesh decimation only has a local effect and can leave part of the meshes unchanged. 
We again train our models on the Faust-Synthetic dataset.
However, since there is no ground-truth correspondence for this version of the dataset, we only present qualitative results using texture transfer from the Faust-Synthetic reference mesh to the test meshes.

We compare MoNet, SplineCNN and FeaStNet on primal meshes to FeaStNet--Dual and our DualConvMax approach on dual meshes in \fig{test_remeshed_train_faust}. 
These texture transfer results show that the correspondence problem for these shapes is substantially more challenging than that for the decimated meshes.
The methods based on the primal mesh fail to recover most correspondences. 
FeaStNet--Dual recovers more correspondences but is overall still very noisy. 
With our DualConvMax approach, we improve the transfer results; see for example the arms. 
This result suggests that our DualConvMax approach learns more robust shape representations that  rely less on the fixed mesh topology of the training meshes. We provide additional qualitative results in the supplementary material.


\begin{table}[t]
    \centering
    \resizebox{\columnwidth}{!}{
    \begin{tabular}{lccccc}
        \toprule
        Input & Trans.\ & Rot.\ & Scale & Geo.\ Err. & Accuracy \\
        \midrule
        XYZ & $\times$ & $\times$ & $\times$ & $1.8$ & $37.3\%$ \\
        Normal & $\checkmark$ & $\times$ & $\checkmark$ & $7.1$ & $40.3\%$ \\
        Area & $\checkmark$ & $\checkmark$ & $\times$ & $27.1$ & $10.6\%$ \\
        Dihedral & $\checkmark$ & $\checkmark$ & $\checkmark$ & $22.6$ & $15.7\%$ \\
        DistCM & $\checkmark$ & $\checkmark$ & $\times$ & $18.1$ & $14.7\%$ \\
        \midrule
        XYZ+Normal & $\times$ & $\times$ & $\times$ & $\textbf{1.3}$ & $45.8\%$ \\
        Normal+DistCM & $\checkmark$ & $\times$ & $\times$ & $2.4$ & $48.3\%$ \\
        Normal+Area+DistCM & $\checkmark$ & $\times$ & $\times$ & $2.5$ & $\textbf{49.2\%}$ \\
        Normal+Dihedral & $\checkmark$ & $\times$ & $\checkmark$ & $6.3$ & $41.3\%$ \\
        Dihedral+Area+DistCM & $\checkmark$ & $\checkmark$ & $\times$ & $13.4$ & $28.9\%$ \\
        \bottomrule
    \end{tabular}
    }
    \caption{Mean geodesic errors and correspondence accuracy  for DualConvMax using different input features  on Faust-Remeshed.
    }
    \label{tab:remeshed_inputs}
\end{table}

\subsection{Results with training on Faust-Remeshed}
\label{ssec:remeshed}

In the experiments so far, none of the methods were exposed to structural changes in the meshes during training on Faust-Synthetic.
In this section, we consider to what extent these  methods can be trained to be robust to  topological changes by training them on the Faust-Remeshed data, where each shape has a unique mesh structure.



\begin{table}
    \centering
    \resizebox{\columnwidth}{!}{
    \begin{tabular}{lllcc}
        \toprule
        Domain   & Method & Input & Geo.\ Er.\  & Acc. \\
        \midrule
        \multirow{3}{*}{Primal} & MoNet \cite{monti17cvpr} & SHOT & $4.1$ & $\textbf{48.7\%}$ \\
         & SplineCNN  \cite{fey18cvpr} & XYZ& $7.2$ & $39.7\%$  \\
         & FeaStNet \cite{verma18cvpr} & XYZ & ${1.6}$ & $47.6\%$ \\
        \midrule
        \multirow{4}{*}{Dual} & \multirow{2}{*}{FeaStNet--Dual}  & XYZ & $1.7$ & $37.8\%$ \\
         & & XYZ+Normal & $1.5$ & $42.4\%$ \\
        \cmidrule{2-5}
         & \multirow{2}{*}{DualConvMax} & XYZ & $1.8$ & $37.3\%$ \\
         & & XYZ+Normal & $\textbf{1.3}$ & $45.8\%$ \\
        \bottomrule
    \end{tabular}}
\caption{Mean geodesic errors and correspondence accuracy on Faust-Remeshed using state-of-the-art methods on primal/dual meshes and using our best performing methods/input features.
}
\label{tab:remeshed_sota}
\end{table}

\mypar{Feature Evaluation}
In \tab{remeshed_inputs} we study the performance of the different input features described in Section~\ref{ssec:features}, as well as  feature combinations, by combining the features based on their particular invariances and performances.


\begin{figure*}
	\resizebox{\textwidth}{!}{
	\begin{tabular}{cccccc}
	 &
	\includegraphics[height=90pt]{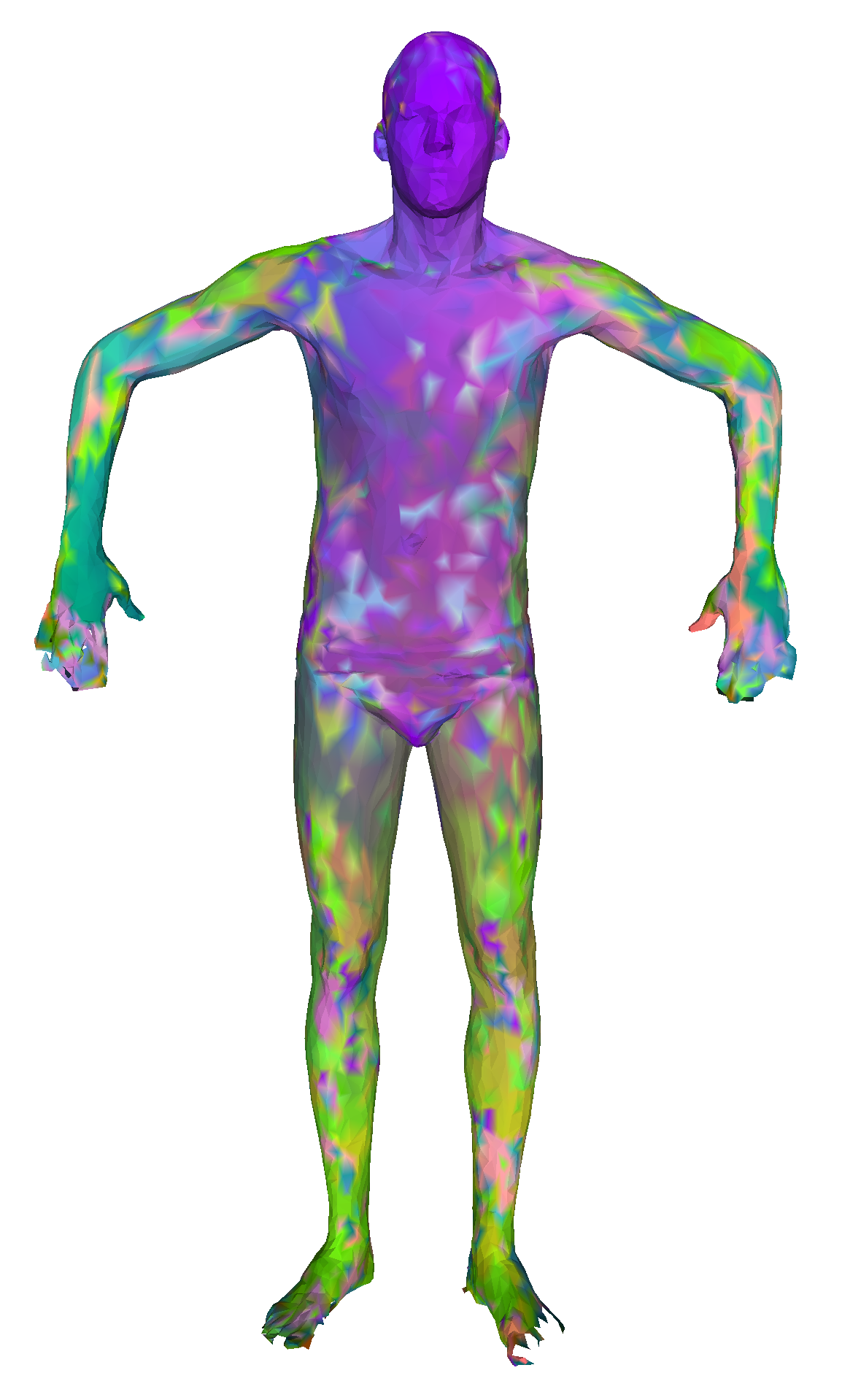} &
	\includegraphics[height=90pt]{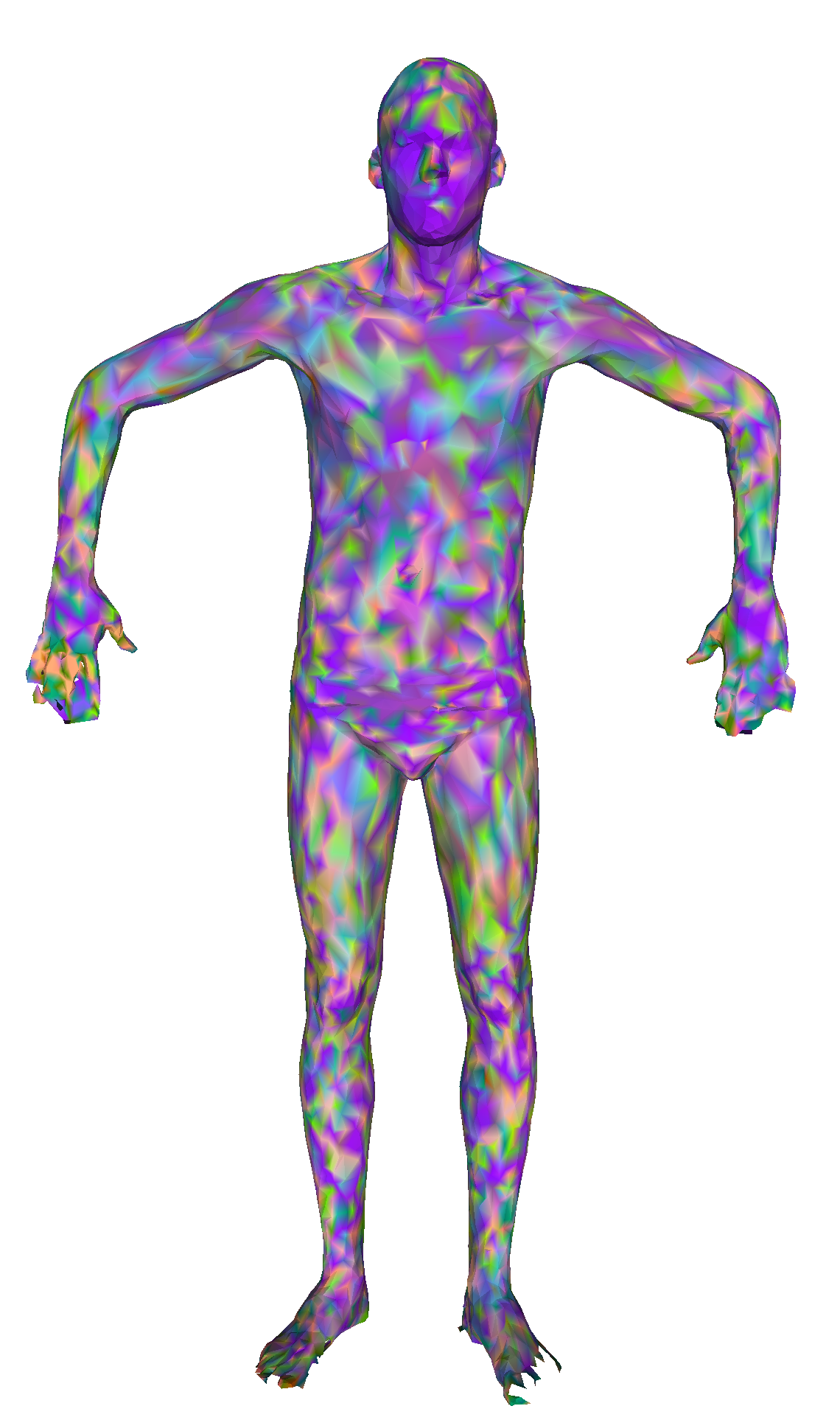} &
	\includegraphics[height=90pt]{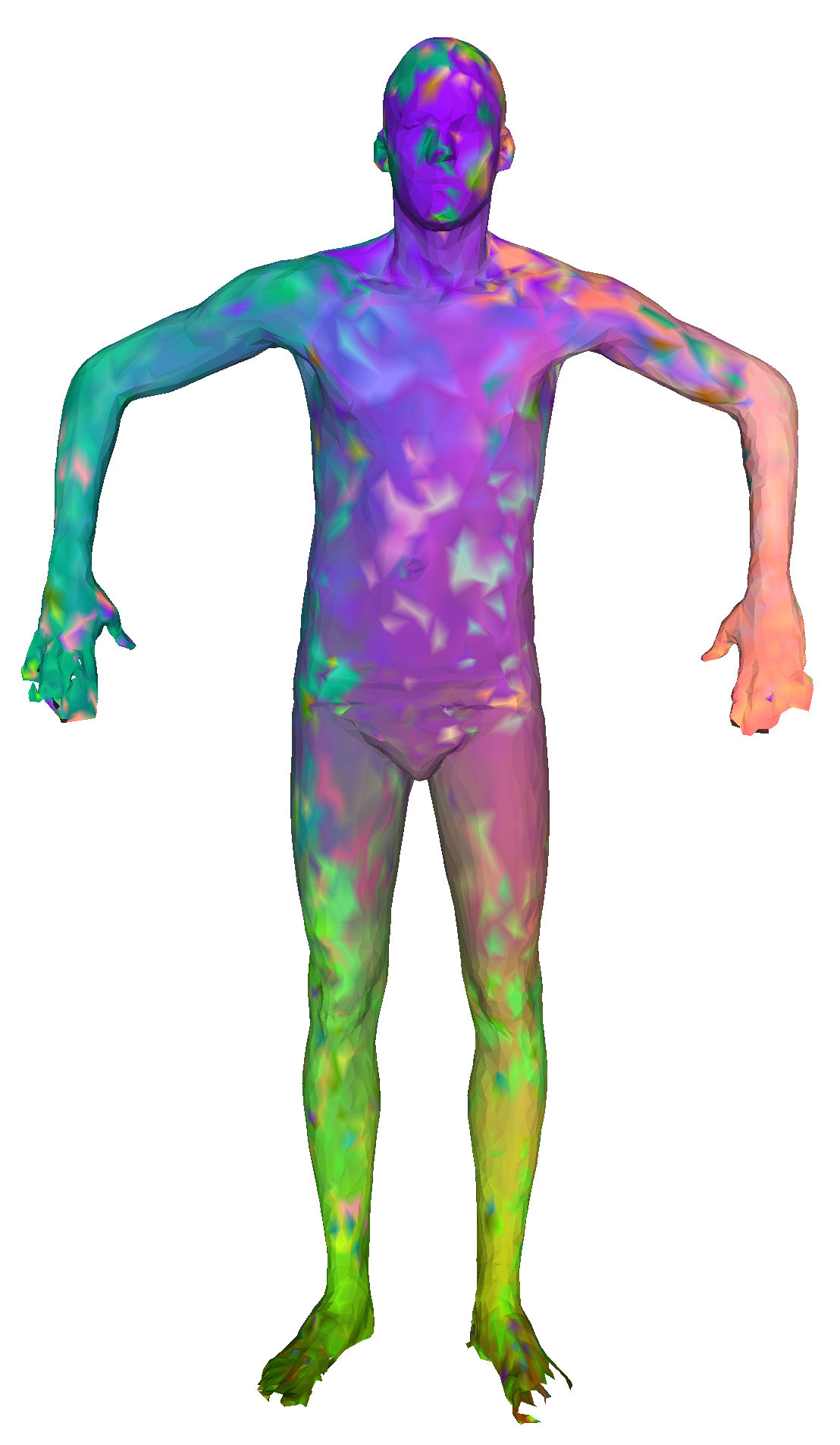} &
	\includegraphics[height=90pt]{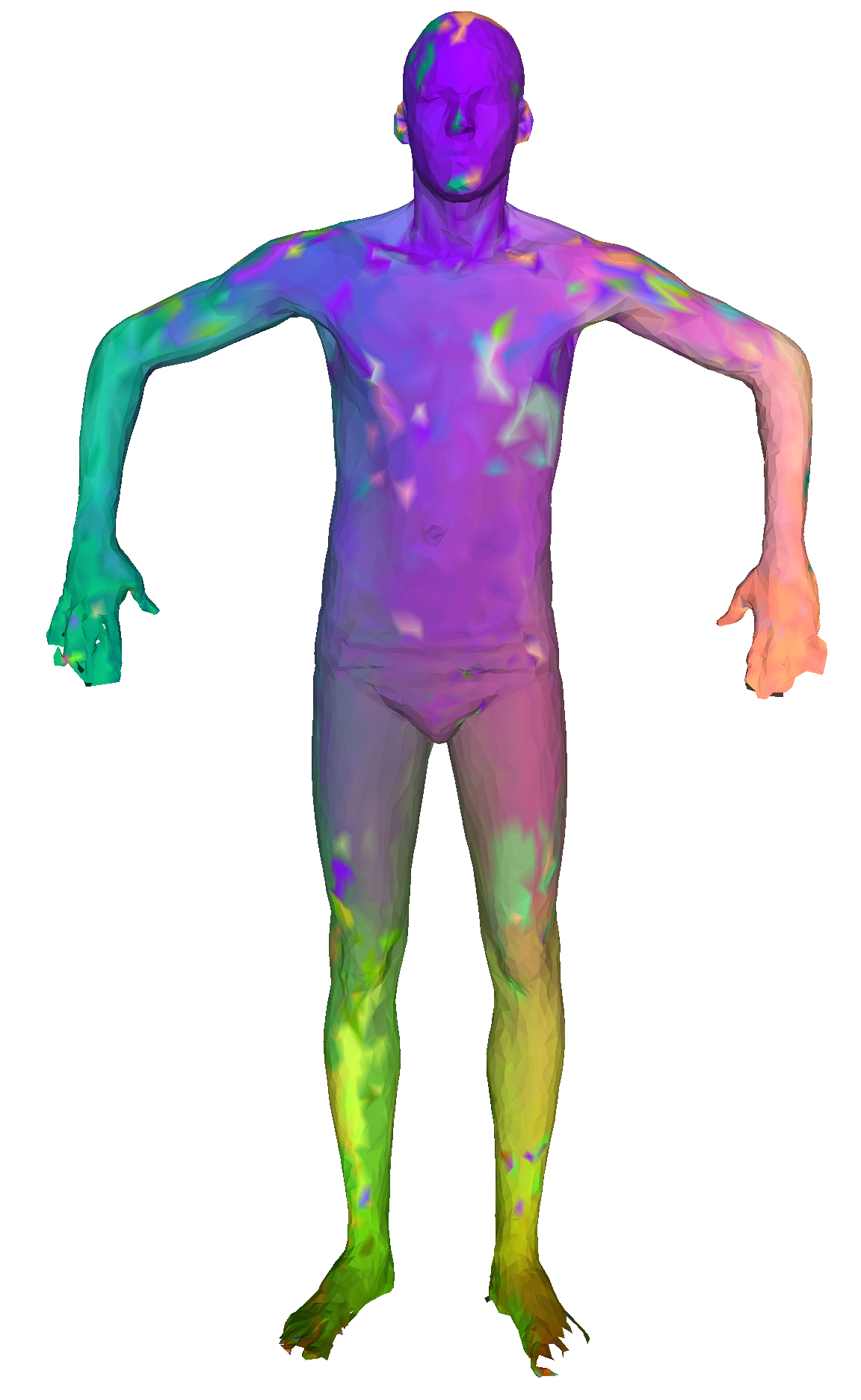} &
	\includegraphics[height=90pt]{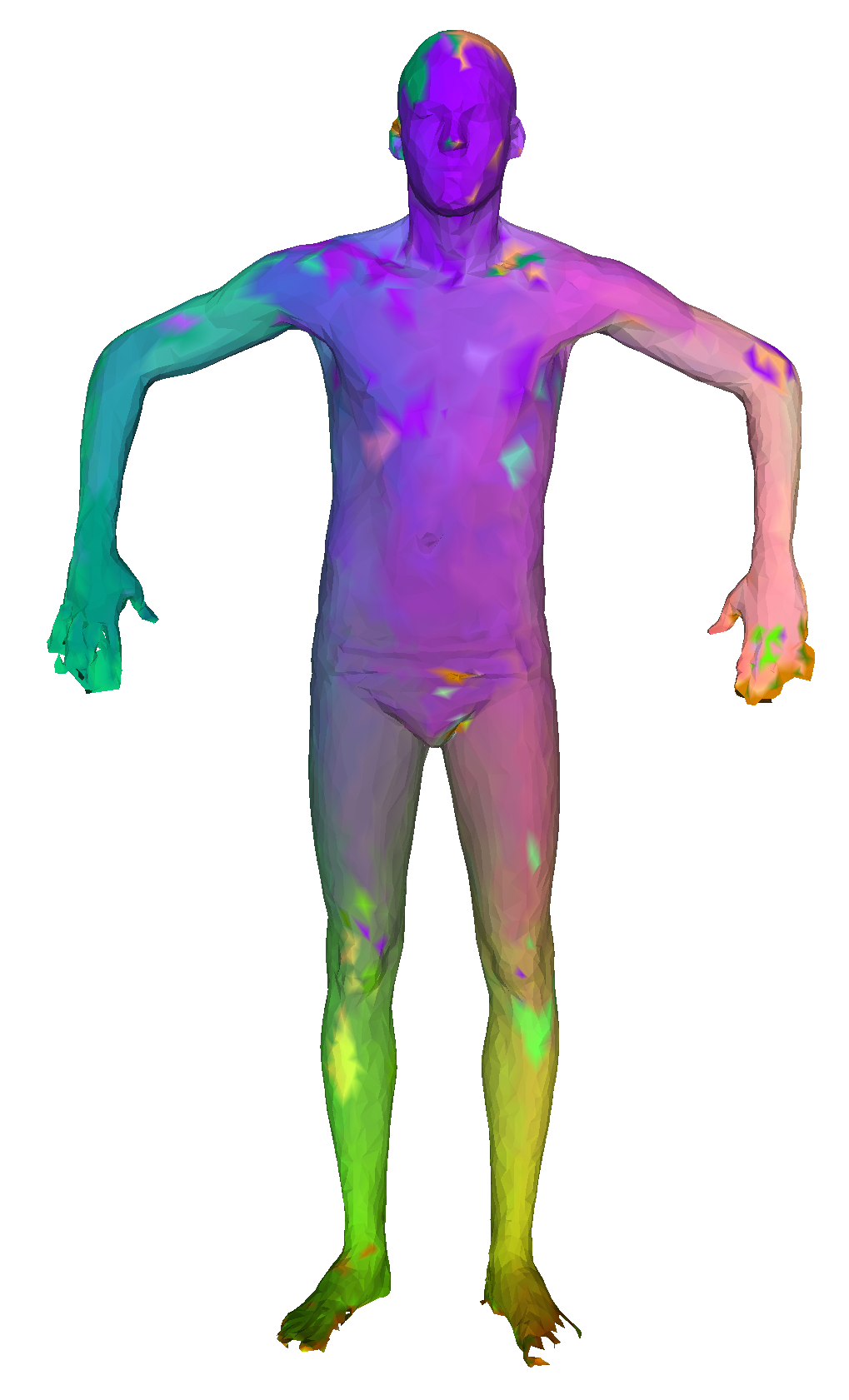} 
	\\
	\includegraphics[height=90pt]{orig_scans/remeshed_faust_ref}
	 &
	\includegraphics[height=90pt]{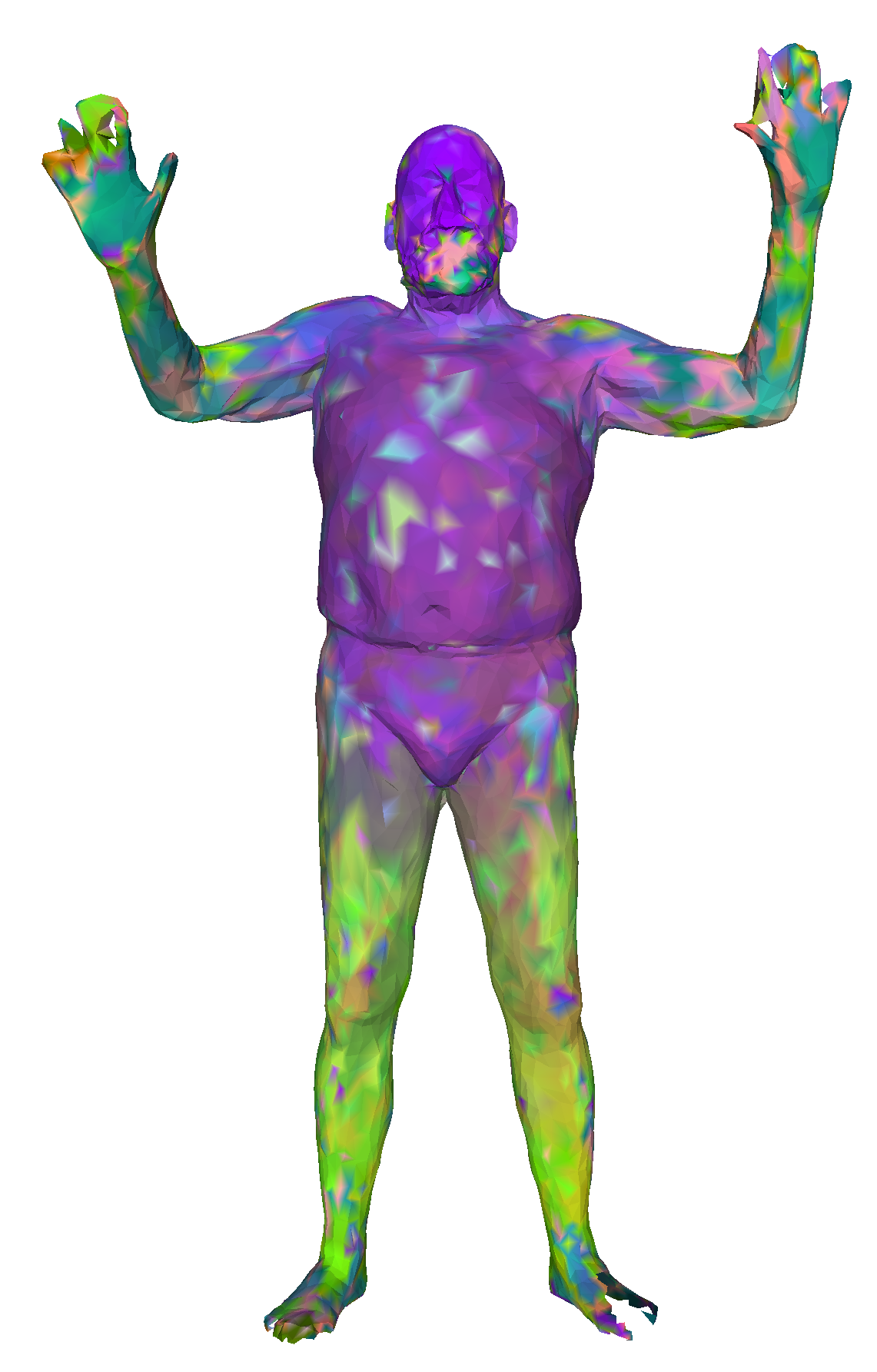} &
	\includegraphics[height=90pt]{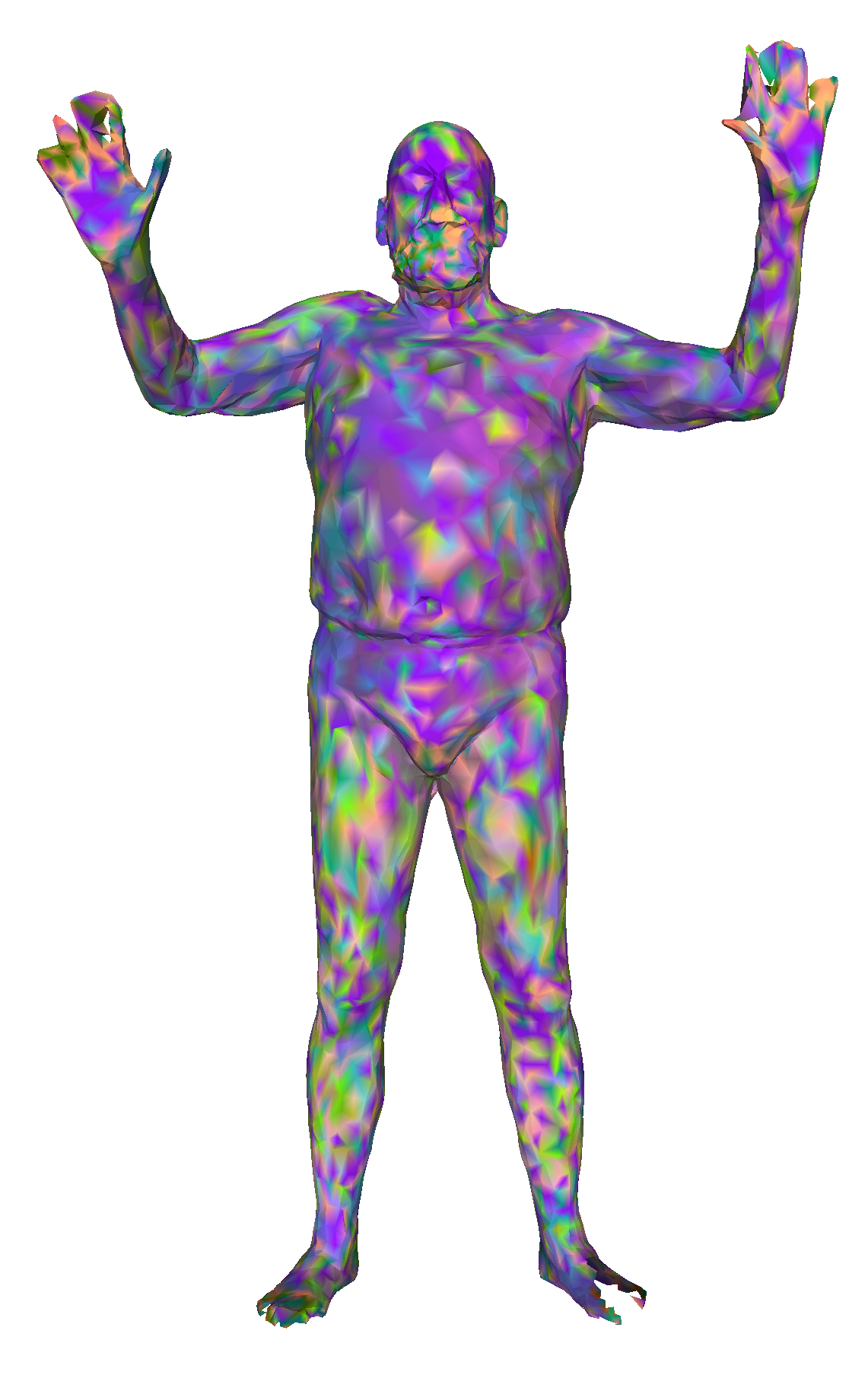} &
	\includegraphics[height=90pt]{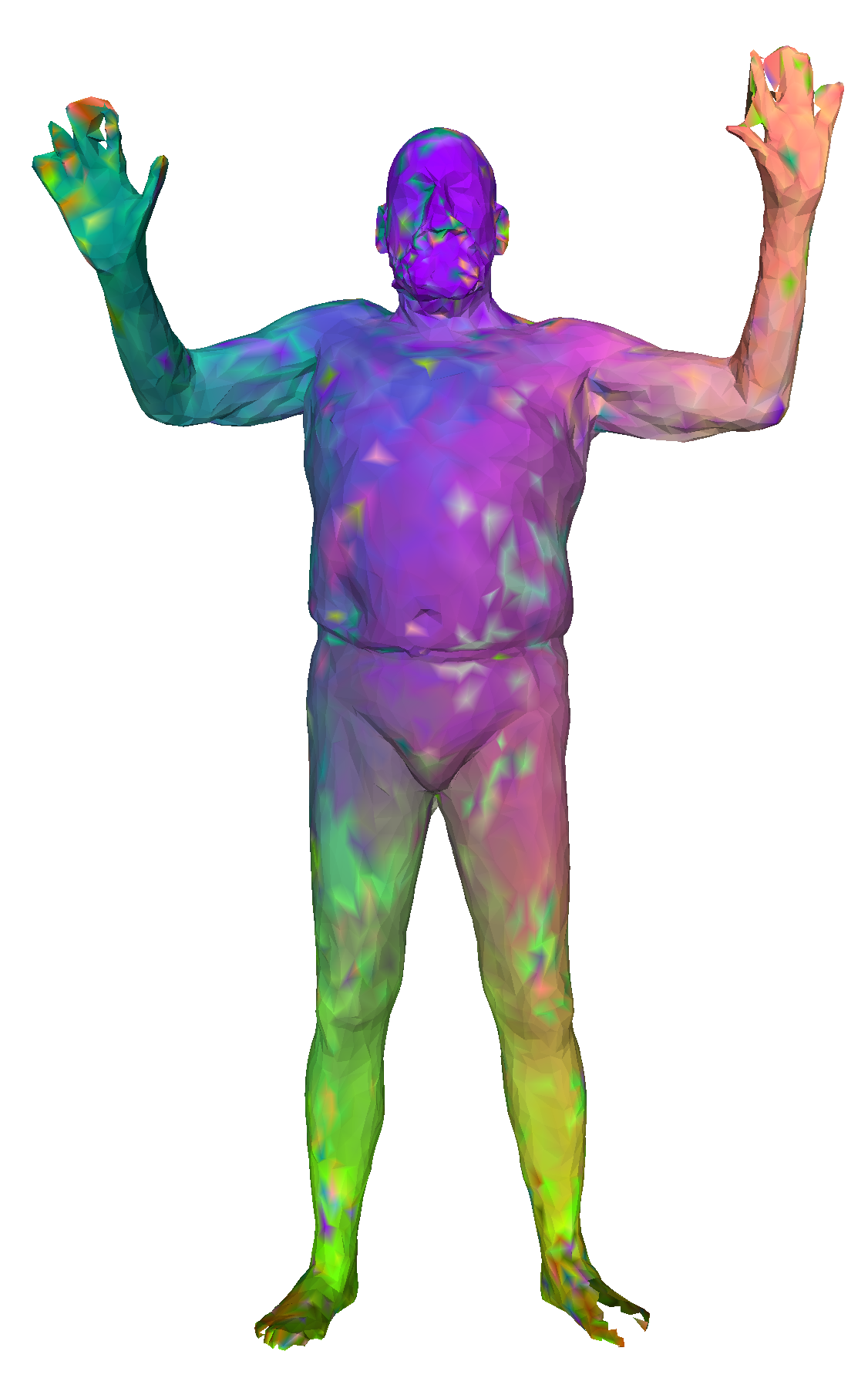} &
	\includegraphics[height=90pt]{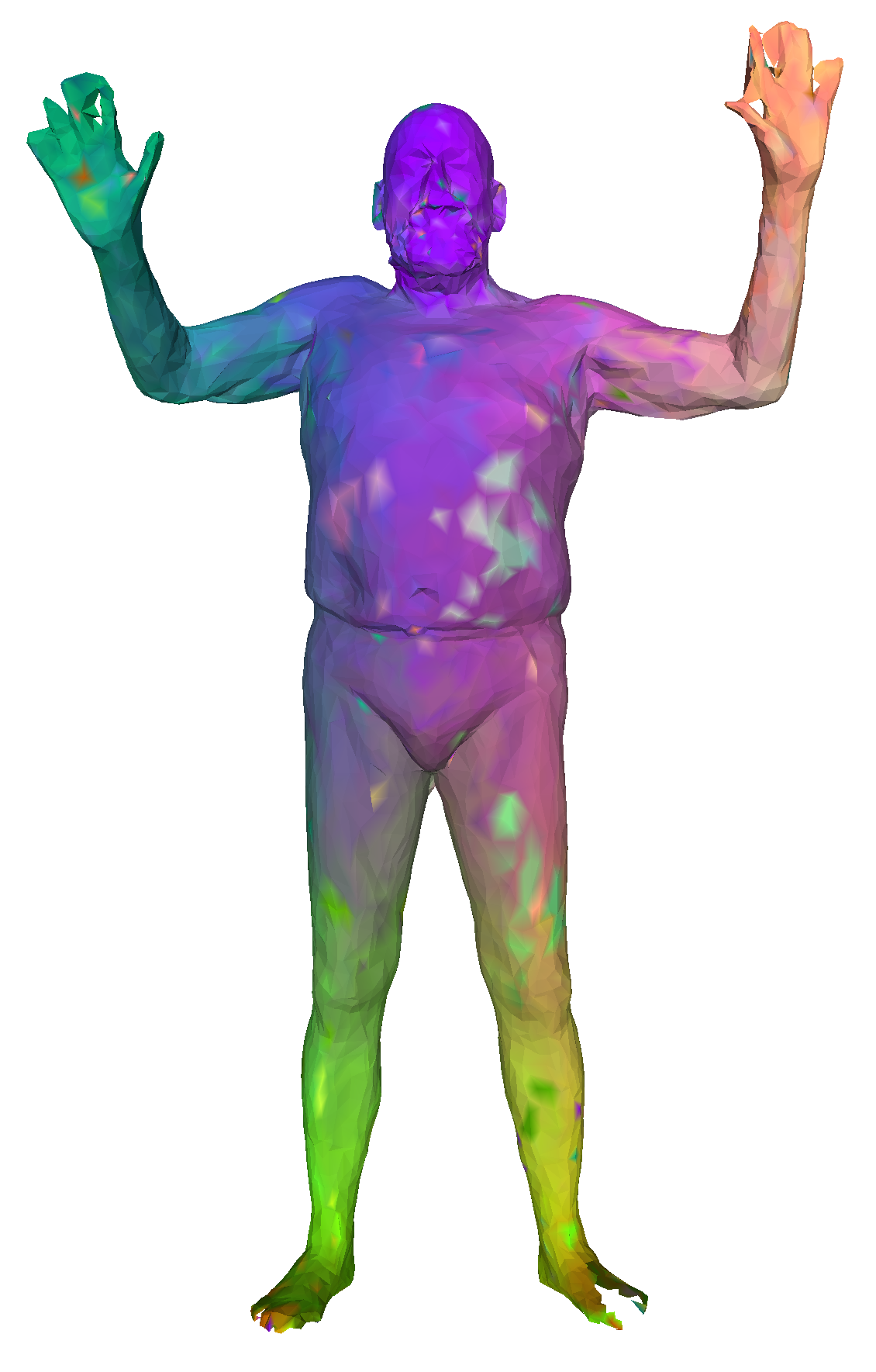} &
	\includegraphics[height=90pt]{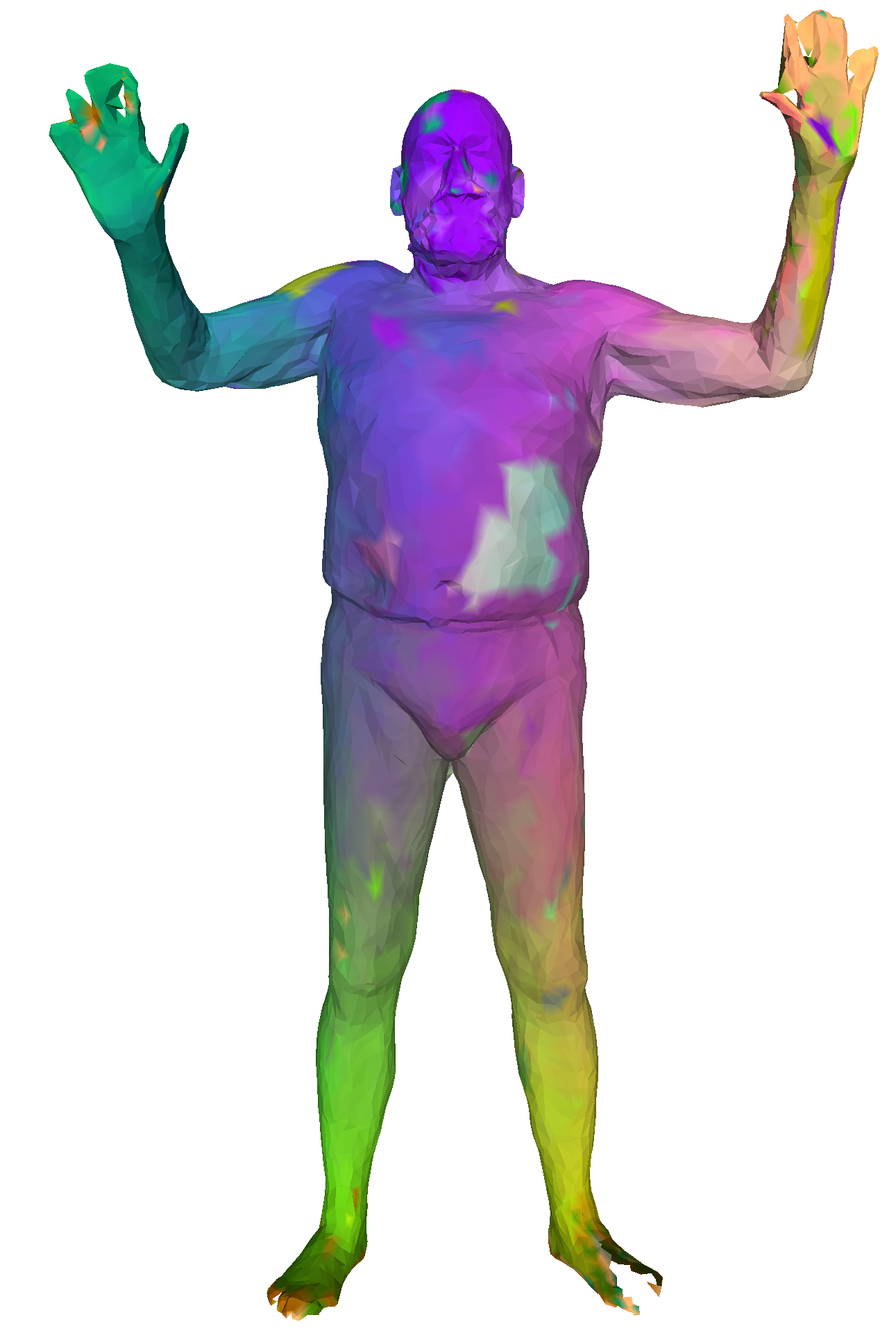} 
	\\
	\footnotesize\textbf{Faust-Remeshed} & \footnotesize\textbf{MoNet} & \footnotesize\textbf{SplineCNN} & \footnotesize\textbf{FeaStNet} & \footnotesize\textbf{FeaStNet--Dual} & \footnotesize\textbf{DualConvMax} 
	\\
	\footnotesize\textbf{Reference} & \footnotesize{(SHOT)} & \footnotesize{(XYZ)} & \footnotesize{(XYZ)} & \footnotesize{(XYZ+Normal)} & \footnotesize{(XYZ+Normal)} 
	\\
	\end{tabular}
}
\caption{Visualizations of texture transfer on Faust-Scan meshes for models trained on the Faust-Remeshed dataset.}
\label{fig:test_scan_train_remeshed}
\end{figure*}

The face normal and XYZ location of the face center provide similar accuracy, well above results obtained using other features. 
While face normals offer translation and scale invariance, the geodesic error is higher as compared to XYZ. 
Combining XYZ and face normals improves over their individual performances and obtains the smallest geodesic error, but does not offer any invariance. 

Among the features which provide translation invariance, we combine Normal and DistCM, which encodes the plane in which the face lies. 
This translation-invariant feature combination yields similar accuracy as the XYZ+Normal combination but yields higher geodesic errors.
To further add translation-invariant face information, we add the area feature.
This achieves the best performance in terms of accuracy,
but yields a minor deterioration in mean geodesic error.
Similarly, we test the combinations Normal+Dihedral and Dihedral+Area+DistCM. Both combinations offer an advantage of translation invariance, plus scale invariance in the former and rotation in the latter, but lead to reduced accuracy and higher geodesic error using the DualConvMax architecture.
We retain the  XYZ feature and the XYZ+Normal feature combination, with the best geodesic error, for the remainder of the experiments.

\mypar{Comparison to previous work} 
In \tab{remeshed_sota}, we compare our  DualConvMax model with previous  state-of-the-art models. 
Among the primal methods, MoNet uses SHOT local shape descriptor features as input, while other models use raw XYZ features.
For the dual methods, we test  XYZ features as well as their combination with face normals. 

Overall, the accuracy and geodesic errors measures on the Faust-Remeshed data are substantially worse than those measured on the Faust-Synthetic data, \cf \tab{faust_dual}.
This underlines the increased level of difficulty of the task on more realistic data.
Among the primal methods, MoNet obtains the highest accuracy (48.7\%), while FeaStNet combines a somewhat lower accuracy (47.6\%) with substantially lower mean geodesic error (1.6 \vs 4.1).
Among the dual methods, DualConvMax with XYZ+Normal features performs best with the best overall mean geodesic error of 1.3 and accuracy (45.8\%) that is comparable but somewhat worse than that of the primal MoNet and FeaStNet.
We provide qualitative evaluations in the supplementary.



\mypar{Qualitative evaluation on Faust-Scan}
Finally, we evaluate all methods trained on the Faust-Remeshed data and visualize texture transfer to the Faust-Scan meshes in \fig{test_scan_train_remeshed}. We observe that training on re-meshed versions of the shapes helps to make primal methods MoNet and FeaStNet more robust to topological changes, \cf \fig{test_scan_train_faust}.
However, we observe that SplineCNN does not  generalize well to topologically different meshes, even after training on the re-meshed data. 
While being more robust to topological changes, the dual-based methods also benefit from training on meshes with varying topology in the Faust-Remeshed dataset. 
When training on Faust-Remeshed, the texture transfer results of dual-based methods are again superior compared to the primal methods, with  DualConvMax yielding the most accurate results overall. For  additional qualitative results see \fig{teaser}.

%% file: sec_conclusion.tex
\section{Conclusion}
\label{sec:conclusion}

We explored the use of the dual mesh to learn shape representations for 3D mesh data as an alternative to the more commonly used primal mesh. 
Performing convolution operations in the dual domain presents the advantage of the neighborhood size being fixed. In addition, it allows access to input features defined naturally on faces, such as normals and face areas. 
We focused our experimental study on the task of real human shape dense correspondence using the Faust human shape dataset. 
We introduced a convolutional operator for the dual mesh and benchmarked it using multiple input features based on the dual mesh. 

In our experiments, we compared our dual mesh approach to existing methods based on the primal mesh and also applied FeaStNet on the dual mesh.
We assess the robustness of different models to topological changes through experiments where we train on one version of the dataset and test on another version of the dataset with different mesh topologies.  
We find that primal methods trained on the Faust-Synthetic dataset, with constant mesh topology across shapes, are brittle and generalize poorly to meshes with different topologies.
This can be remedied to some extent by training on meshes with varying topology, as we did using the Faust-Remeshed dataset.
Our results show the robustness of our convolutional operator applied on the dual mesh by achieving the best performances when testing structurally different meshes, whether they are trained on fixed or variable mesh structures.


Although we focused on shape correspondence in the current paper, it is interesting to explore in future work the use of the dual mesh to define deep networks for other tasks such as shape matching, classification, and semantic segmentation of meshes.
